\documentclass[11pt]{article}

\usepackage[final]{acl}

\usepackage{times}
\usepackage{latexsym}

\usepackage[T1]{fontenc}

\usepackage[utf8]{inputenc}

\usepackage{microtype}

\usepackage{inconsolata}

\usepackage{graphicx}

\usepackage[table]{xcolor}
\usepackage{caption}
\usepackage{fvextra}
\usepackage{subcaption}
\usepackage{wrapfig}
\usepackage{booktabs}
\usepackage{placeins}

\usepackage[most]{tcolorbox}
\tcbset{
  colback=blue!3!white,
  colframe=blue!60!black,
  boxrule=0.8pt,
  arc=4pt,
  left=6pt, right=6pt, top=6pt, bottom=6pt
}
\tcbuselibrary{breakable}

\usepackage{tabularx}
\usepackage{pifont}

\newcommand{\cmark}{\textcolor{green!60!black}{\ding{51}}}
\newcommand{\xmark}{\textcolor{red}{\ding{55}}}

\usepackage{dsfont}
\usepackage{pifont}
\usepackage[table]{xcolor}

\usepackage{amsfonts}
\usepackage{makecell}
\usepackage{mathtools}
\usepackage{enumitem} 

\usepackage{multirow}
\usepackage{booktabs}
\usepackage{makecell}
\usepackage{graphicx}

\title{Are We Measuring Strategy or Phrasing?\\The Gap Between Surface- and Approach-Level Diversity\\in LLM Math Reasoning}

\author{
  Sangmook Lee$^{1}$, Minbeom Kim$^2$, Jeonghye Kim$^3$, \\
  {\bfseries Dohyung Kim$^1$, Sojeong Rhee$^3$, Kyomin Jung$^{1 \dagger}$ } \\[5pt]
  $^1$Seoul National University, $^2$SK hynix, $^3$KAIST
  \\
  \texttt{\{helmsman, kjung\}@snu.ac.kr}
}

\begin{document}
\maketitle
\renewcommand{\thefootnote}{\fnsymbol{footnote}}
\footnotetext[2]{Corresponding author.}
\renewcommand{\thefootnote}{\arabic{footnote}}

\begin{abstract}

Diversity in LLM mathematical reasoning is critical for exploration, but common diversity metrics mostly capture \textit{surface-level variation} rather than differences in how a problem is solved. We address this gap by introducing \textbf{approach-level diversity}: variation in strategies across correct solutions to the same problem. Using a human-calibrated LLM judge framework, we show that prior diversity measures are unreliable proxies for approach-level diversity, and this mismatch carries over to diversity-aware RLVR, where target metrics are preserved while approach-level diversity declines. Investigating when approach-level diversity helps and whether it can be directly induced, we find that approach-diverse candidate sets improve test-time scaling. However, optimizing an LLM judge diversity reward during training causes the policy to exploit judge-specific preferences rather than broaden its approaches, leaving direct optimization of approach-level diversity as an open problem. Together, our work introduces the notion of approach-level diversity and uncovers a systematic divergence between surface- and approach-level signals, marking a step toward LLMs that reason in genuinely diverse, human-like ways.
Our code and data are publicly available at \url{https://github.com/helmsman12/LLM_reasoning_diversity}.

\end{abstract}

\section{Introduction}

Diversity in large language models (LLM) generations has become practically important across tasks where useful outputs span a range of plausible alternatives, such as creative writing \citep{creativewriting} and synthetic data construction \citep{shumailov2024ai}. In complex reasoning tasks, this importance is amplified by test-time scaling methods such as verifier-based selection \citep{lightman2023letsverifystepstep} and multi-agent collaboration \citep{wu2024autogen}, where downstream performance depends on the range of candidate solutions sampled from the model.

However, recent studies report that reinforcement learning with verifiable rewards (RLVR), a dominant paradigm for post-training language models, reduces the diversity of a trained policy \citep{kirkunderstanding, he2025rewarding, yuedoes, wu2025invisible}, prompting diversity-aware RLVR methods that attempt to preserve diversity during training.
Yet we find that these methods typically operationalize diversity by \textit{surface-level diversity}---differences in the observable form of outputs---through measures such as lexical overlap \citep{lanchantin2025diverse}, embedding distance \citep{chen2026dqo}, or symbolic representations like the ratio of unique equations \citep{hu2026diver}.
Such measures reveal whether solutions differ in wording or representation, but leave open a more fundamental question: \textit{are models producing surface-level variants of the same strategy, or exploring genuinely different ways to solve the problem?}

\begin{figure*}[t]
  \centering
  \includegraphics[width=\textwidth]{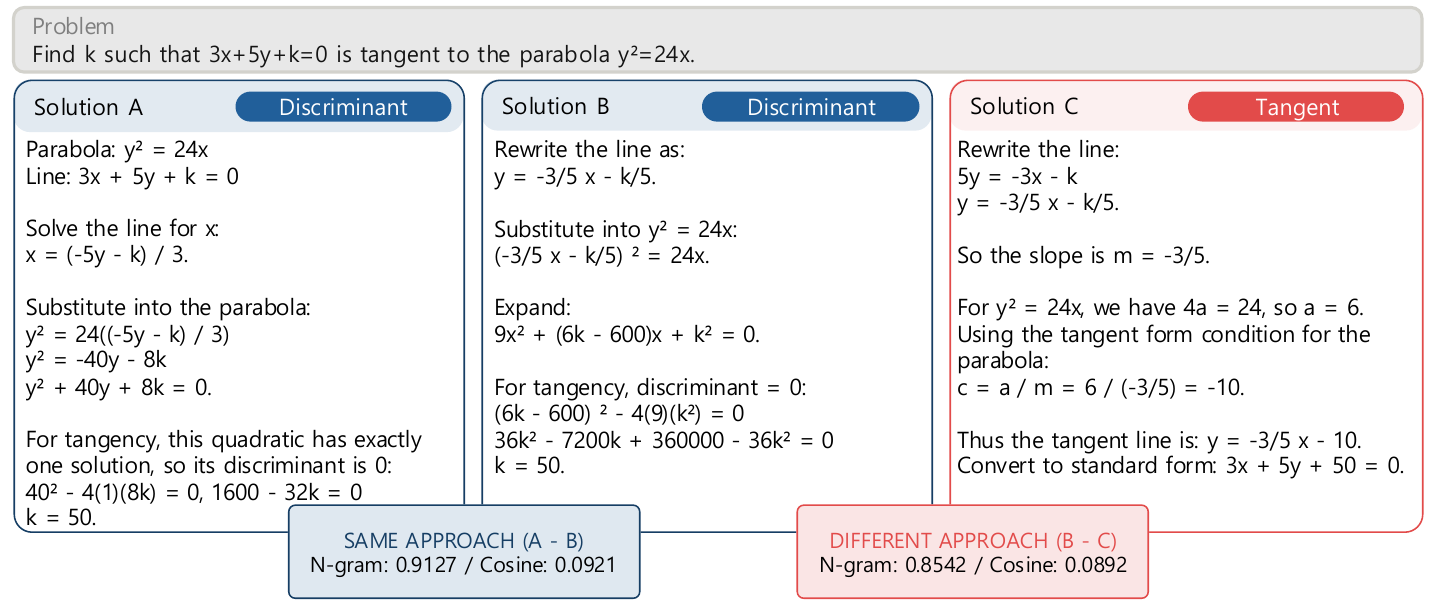}
  \caption{\textbf{A real-world failure case of conventional diversity metrics.}
Conventional metrics assign a higher diversity score to a pair of solutions that follow the same approach than to a pair that uses different mathematical approaches, illustrating a mismatch between surface-level variation and approach-level diversity.}
  \label{fig:fail-case}
\end{figure*}

To address this gap, we introduce \textbf{approach-level diversity} as a distinct axis of LLM reasoning behavior: variation in the underlying solution strategies used to arrive at the correct answer, beyond differences in wording, notation, or exposition.
Prior work measures diversity at two common levels: lexical, typically via N-gram distance, and semantic, typically via cosine similarity over embeddings. We first evaluate whether these standard metrics align with human judgments of approach-level diversity.
We find that these metrics are susceptible to paraphrastic variation in math reasoning, where changes in wording, layout, or symbolic expression can inflate the apparent diversity among solutions with the same reasoning approach.

Building on these observations, we design a scalable evaluation framework for approach-level analysis built around a human-calibrated LLM judge. Using this framework, we extend the analysis to a wider range of prior measures, including math-domain-specific diversity metrics, and show that they remain unreliable for detecting fine-grained differences in approaches. 
This mismatch between surface- and approach-level diversity persists in recent diversity-aware RLVR algorithms. Although these methods preserve the diversity proxies they optimize, this does not imply that approach-level diversity is preserved.
In practice, training often increases surface variation within a narrower set of approaches, making outputs look more diverse even as approach-level diversity declines.

We finally ask two questions: what practical benefits approach-level diversity can provide, and whether an LLM judge approach-diversity signal that works for evaluation can also be used as a training reward. 
Our results reveal a gap between measuring and optimizing approach-level diversity. 
Candidate sets containing distinct approaches yield larger gains under test-time scaling, suggesting that approach-level diversity can improve inference-time performance.
However, when directly optimized for the LLM judge approach-level diversity signal, the policy learns to satisfy the judge by exploiting its bias rather than generating genuinely diverse approaches, calling for a training-compatible, robust approach-level diversity metric.

Together, our work introduces approach-level diversity as a distinct axis for analyzing mathematical reasoning in LLMs, along with a human-calibrated framework for measuring it. Using this framework, we uncover a systematic gap between surface-level and approach-level diversity: existing metrics and diversity-aware training methods capture how solutions are written rather than how problems are solved. By discovering this gap, our work marks a step toward developing LLMs that exhibit greater strategic flexibility in mathematical reasoning.


\section{Background}

\subsection{Related Work}

\paragraph{Diversity collapse in RL post-training}
RL-based post-training has substantially improved the generation quality of LLMs \citep{schulman2017ppo, shao2024deepseekmath}.
Recent work, however, reports that RL post-training can reduce generation diversity, leading to narrower task coverage \citep{wu2025invisible, yuedoes, kirkunderstanding, cuientropy} and more homogeneous outputs \citep{chen2026dqo, hu2026diver}.
This has motivated diversity-aware RLVR methods that aim to preserve diversity during training by adding auxiliary rewards or regularizers based on likelihood \citep{he2025rewarding, lanchantin2025diverse}, embedding-space distance \citep{chen2026dqo, chen2025dra}, or token-level overlap \citep{hu2026diver, lanchantin2025diverse}.
Because most of these works focus on math reasoning tasks—a dominant testbed for RLVR methods—several have proposed diversity measures for mathematical reasoning to analyze such diversity collapse and guide mitigation efforts.

\paragraph{Measuring diversity in mathematical reasoning}
Diversity in mathematical reasoning has been measured with lexical overlap \citep{hu2026diver}, symbolic proxies \citep{wuprogress, he2025rewarding}, embedding similarity \citep{chen2026dqo}, and LLM-based reasoning-path decompositions \citep{ju2026reasoningpathdivergencenew}.
However, these metrics define difference through their chosen representation, leaving open whether they align with human judgments of distinct solution strategies.
Although human-aligned diversity has been studied in open-ended generation \citep{jiang2025artificialhivemindopenendedhomogeneity, zhang2025noveltybench}, it remains less explored in mathematical reasoning.
Existing LLM-judge-based evaluations in this domain \citep{li2025jointly, ju2026reasoningpathdivergencenew, zhu2026flowrl} do not explicitly define the notion of diversity they measure or validate judge decisions against human judgments.
We address this gap by formalizing approach-level diversity as strategy-level variation among correct solutions and by testing whether existing diversity measures and diversity-aware RLVR methods align with this definition.

\subsection{Diversity Metrics under Evaluation}
\label{sec:metrics_under_eval}

We summarize the diversity metrics serving as baselines in our evaluation.
We evaluate five diversity metrics covering lexical, semantic, symbolic, and reasoning-level signals. N-gram distance \citep{kondrak2005n} and Self-BLEU \citep{zhu2018texygen} measure token-level lexical overlap, while cosine distance over Qwen3-Embedding-8B \citep{qwen3embedding} measures semantic similarity.
We also include two math-specific metrics: Distinct-Equations \citep{wuprogress}, which measures the ratio of unique equations in a solution set, and RPD \citep{ju2026reasoningpathdivergencenew}, which decomposes solutions into reasoning steps and aggregates step-level embedding distances.
Formal definitions are in Appendix~\ref{sec:div_met_def}.

\begin{table}[t]
\centering
\label{tab:metric-comparison}
\small
\setlength{\tabcolsep}{3.5pt}
\renewcommand{\arraystretch}{1.10}
\begin{tabular}{@{}l l c c@{}}
\toprule
Metric & Signal & \makecell{Model-\\based} & \makecell{Structure-\\aware} \\
\midrule
N-gram        & Lexical    & \xmark & \xmark \\
Self-BLEU     & Lexical    & \xmark & \xmark \\
Cosine        & Semantic   & \cmark & \xmark \\
Distinct-EQ   & Symbolic   & \xmark & \cmark \\
RPD           & Reasoning  & \cmark & \cmark \\
\bottomrule
\end{tabular}
\caption{Taxonomy of diversity metrics analyzed.}
\end{table}

\section{Conventional Metrics Do Not Capture Approach-level Diversity}

We begin by defining the object of study: approach-level diversity in closed-ended mathematical reasoning.
We collect human reference labels for this notion and use them to diagnose where representative conventional metrics fail, tracing these failures to structural properties of mathematical solutions.

\subsection{Defining Approach-level Diversity}
\label{sec:def_reasoning_diversity}
We define \textbf{\textit{approach-level diversity}} as the variation in the underlying problem-solving strategies used to solve the same mathematical problem correctly.
This is distinct from \textbf{\textit{surface-level diversity}}, which refers to the variation in the observable form of a solution, such as wording, notation, formatting, equation layout, or exposition style.
Two solutions can look different while following the same strategy, or share much of their observable form while relying on different mathematical mechanisms.
Approach-level diversity captures the latter distinction, which we operationalize as follows.

\begin{center}
\begin{tcolorbox}[
    width=0.96\linewidth,
    colback=gray!5,
    colframe=black!60,
    boxrule=0.5pt,
    arc=3pt,
    left=6pt, right=6pt, top=6pt, bottom=6pt,
    title=\textbf{\small Definition of Different Approaches},
    fonttitle=\bfseries,
  ]
  \small
  We say two solutions follow different approaches if they diverge meaningfully along at least one of the following three dimensions:
  \begin{itemize}[leftmargin=1.2em, topsep=2pt, itemsep=1pt, parsep=0pt]
      \item \textbf{Mathematical tools} --- the techniques invoked
            (e.g., algebraic vs.\ geometric).
      \item \textbf{Structural definitions} --- how the problem is set up
            (e.g., auxiliary function vs.\ direct substitution).
      \item \textbf{Representational viewpoint} --- the perspective taken
            (e.g., coordinate vs.\ synthetic geometry).
  \end{itemize}
\end{tcolorbox}
\end{center}

\subsection{Human Judgments of Approach Diversity}
\label{sec:human_anno}

To test whether conventional diversity metrics capture approach-level diversity, we construct a human reference for metric comparisons. Following Section~\ref{sec:def_reasoning_diversity}, annotators judge whether two correct solutions to the same problem use the same underlying approach.
Our annotation set contains 80 solution pairs from 20 problems from the MATH \citep{hendrycksmath} training set, with four model-generated pairs per problem.
Each item includes the problem and two correct model-generated solutions, and receives two independent labels from 17 annotators. All annotators hold at least a bachelor's degree in an engineering-related field, a requirement we impose to ensure sufficient familiarity with mathematical reasoning.

Annotators agreed on 80\% of the items, suggesting that the definition can be applied with reasonable consistency. 
The disagreements were resolved by the authors. After this, we use the resulting labels as the human reference for the analyses below. Further details are provided in Appendix~\ref{app:human_annotation}.




\begin{figure*}[t]
    \centering

    \begin{subfigure}[t]{0.36\textwidth}
        \centering
        \includegraphics[width=\linewidth]{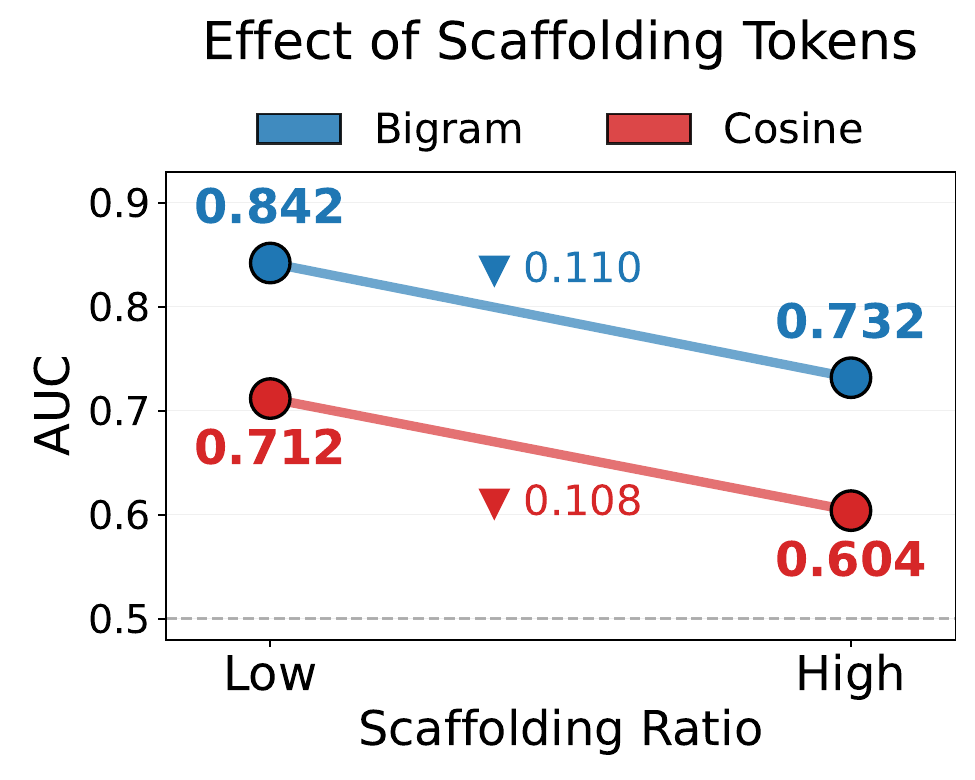}
        \caption{Scaffolding tokens}
        \label{fig:scaffolding_ablation}
    \end{subfigure}
    \hfill
    \begin{subfigure}[t]{0.62\textwidth}
        \centering
        \includegraphics[width=\linewidth]{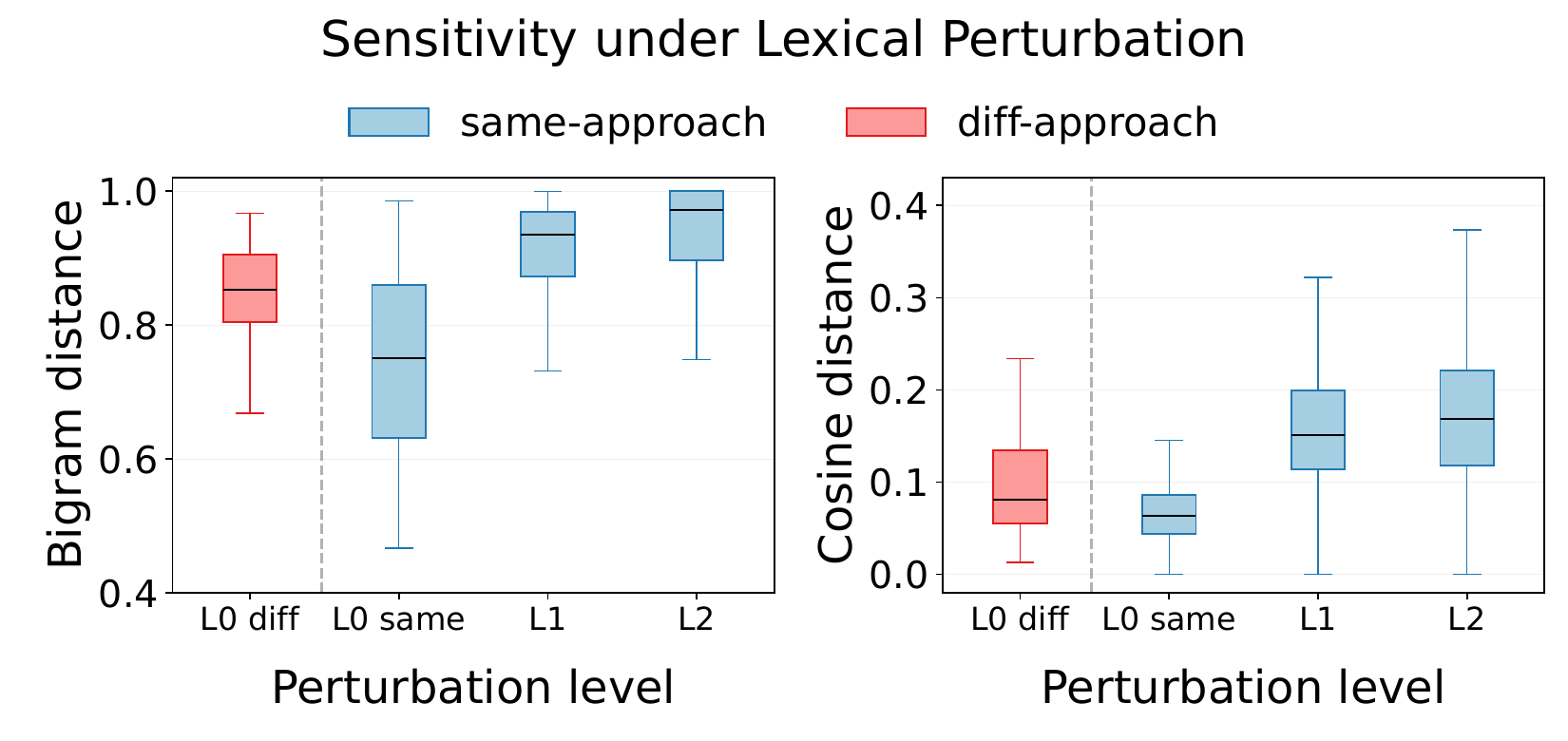}
        \caption{Paraphrase perturbation}
        \label{fig:paraphrase_ablation}
    \end{subfigure}

    \caption{\textbf{Limitations of conventional diversity metrics.}
    (a) Conventional metrics fail to discriminate approach signals under heavy scaffolding.
    (b) Conventional metrics remain sensitive to realistic paraphrases.}
    \label{fig:paraphrase-scaffolding}
\end{figure*}

\subsection{Failure Modes of Conventional Metrics}
\label{sec:fail_model_cov_measures}
We compare the human labels with two representative conventional metrics: bigram distance and cosine distance. 
The metrics disagree with human judgments in both directions: they sometimes assign high distances to same-approach pairs and low distances to different-approach pairs. 
We trace this mismatch to two properties of solutions in mathematical reasoning.

\paragraph{Two Confounds Behind the Disagreement}
The disagreement stems from two factors that conventional metrics cannot
separate from the actual approach.
The first is \textbf{\textit{shared scaffolding}}: content that most correct solutions to a problem include regardless of approach, such as restated conditions, common setup steps, and answer formatting. The second is \textbf{\textit{approach-preserving paraphrasing}}: changes in wording, notation, equation arrangement, or exposition that alter how a solution looks while leaving its underlying approach unchanged.

Figure~\ref{fig:fail-case} illustrates a real-world example. Solutions A and B use the same discriminant-based approach but differ in local algebra, while Solutions B and C use different mechanisms but share similar notation.
Thus, the metrics assign higher diversity to the same-approach pair than to the different-approach pair, reversing the human decision. 

\paragraph{Scaffolding Weakens Approach Signals}
When solutions share substantial scaffolding, conventional metrics may struggle to isolate approach-specific signals. We test this by asking whether their discriminative ability degrades on high-scaffolding problems.
For each problem in our human annotation set, we compute a scaffolding score as the ratio of overlapping unigrams among correct solutions, and split problems at the median into low- and high-overlap groups. The median overlap is 61.2\%, suggesting that shared setup accounts for a large portion of solutions.
As Figure~\ref{fig:scaffolding_ablation} shows, both metrics become substantially less discriminative in the high-overlap group.
This indicates that shared scaffolding can dominate solution-level similarity scores, masking the smaller portion that reflects the approach. Details of the scaffolding-ratio computation are provided in Appendix~\ref{app:scaffolding_ratio}.

\paragraph{Sensitivity to Paraphrasing}
Even pure rewording can prevent conventional metrics from measuring approach-specific differences.
To test this, we generate approach-preserving paraphrases at two levels.
Starting from each original solution (L0), L1 applies local wording or notation rewrites (e.g., $y=mx+b$ as $mx-y+b=0$), while L2 additionally changes the global presentation format (e.g., verbose versus compact exposition);
we generate three paraphrases for each original solution.

We then compare the distance distributions of rewritten same-approach solution pairs with those of original different-approach pairs.
Figure~\ref{fig:paraphrase_ablation} shows that even L1 rewrites are enough to 
push same-approach pairs farther apart than different-approach pairs, 
and the small gap between L1 and L2 indicates that most of this 
distortion already arises from local rewording rather than presentation 
shifts. 
Thus, conventional metrics are vulnerable even to realistic paraphrases, reflecting how a solution is written rather than how it is solved.

\section{Approach Mismatch Persists at Scale}

The previous section showed that conventional metrics cannot reliably 
capture approach-level diversity, motivating an evaluation framework 
that looks beyond surface form. 
We build such a framework around an LLM judge and use it to scale our analysis to real model generations, broadening the set of metrics under study and applying the framework to recent diversity-aware RLVR methods.

\subsection{Approach Evaluation with LLM Judge}
\label{sec:approach_judge_design}

In this section, we describe our LLM-judge-based evaluation framework and explain the design choices behind it.
Implementation details are provided in Appendix~\ref{app:details_eval_framework} and ~\ref{app:prompts}.
To build a robust and reliable evaluation framework, we consider the following requirements.

\begin{itemize}[leftmargin=*, itemsep=2pt, topsep=3pt]
\item \textbf{D1: Approach-feasible evaluation.} Approach-level diversity can be meaningfully evaluated only on problems that admit multiple approaches.
\item \textbf{D2: Scalable and interpretable labeling.} Evaluation should scale across many problems and models while producing labels that can be reused in downstream analyses.
\item \textbf{D3: Human-calibrated decision boundary.} The judge should align with human judgments, especially on clear cases where annotators agree.
\end{itemize}

\noindent We address these three requirements as follows.

\paragraph{Approach-feasible problem set}
\label{sec:multi_approach_problems}
First, we filter for problems that admit multiple correct approaches. 
For each candidate problem, we (i) prompt GPT-5.2 \citep{gpt_52} to generate four distinct approaches, (ii) ask Qwen3-4B to solve the problem with each one and keep only those that reach the correct answer, and (iii) use a GPT-5.2 judge to confirm that the surviving approaches are mutually distinct. We retain only problems with at least three unique and correct approaches, giving 469 problems sampled from the MATH training set \citep{hendrycksmath}.
The remaining analyses in this section are performed on this evaluation set. The cost report for the pipeline can be found in Appendix~\ref{app:details_eval_framework}.

\paragraph{Clustering with an LLM judge}
\label{sec:llm_diversity_judge}

To obtain scalable and reusable approach labels, we ask the judge to cluster correct solutions for each problem by their underlying approach.
Given a problem $x$ and a set of correct solutions $S_x$, an LLM Judge $\mathcal{J}$ powered by GPT-5.2 partitions $S_x$ into approach-level clusters,
\begin{equation}
\mathcal{J}(x, S_x) = {C_1, \ldots, C_{K_x}}, \quad S_x = \bigsqcup_{i=1}^{K_x} C_i.
\label{eq:clustering_equation}
\end{equation}

\noindent Each cluster $C_i$ contains solutions judged to follow the same approach, and $K_x$ denotes the number of distinct approaches identified for problem $x$.

\paragraph{Agreement with human annotations}
We validate the judge against the human labels from Section~\ref{sec:human_anno}.
The judge achieves 85.0\% agreement with the human reference labels, comparable to the 80.0\% inter-annotator agreement.
Moreover, among the 64 clear cases on which all annotators agreed, the judge disagreed on only 2, indicating that the judge aligns well with human judgments.

\medskip

\noindent
We adopt GPT-5.2 as the main backbone model due to its high reasoning and instruction-following ability. 
To account for a possible judge-specific bias, we re-ran all judge-dependent stages of our pipeline, such as problem filtering and approach classification, with an independent Claude Sonnet 5 \citep{claude_sonnet_5} judge. As shown in Appendix~\ref{app:details_eval_framework}, all conclusions in this paper hold under either judge.

\begin{figure*}[t]
\centering
\includegraphics[width=\textwidth]{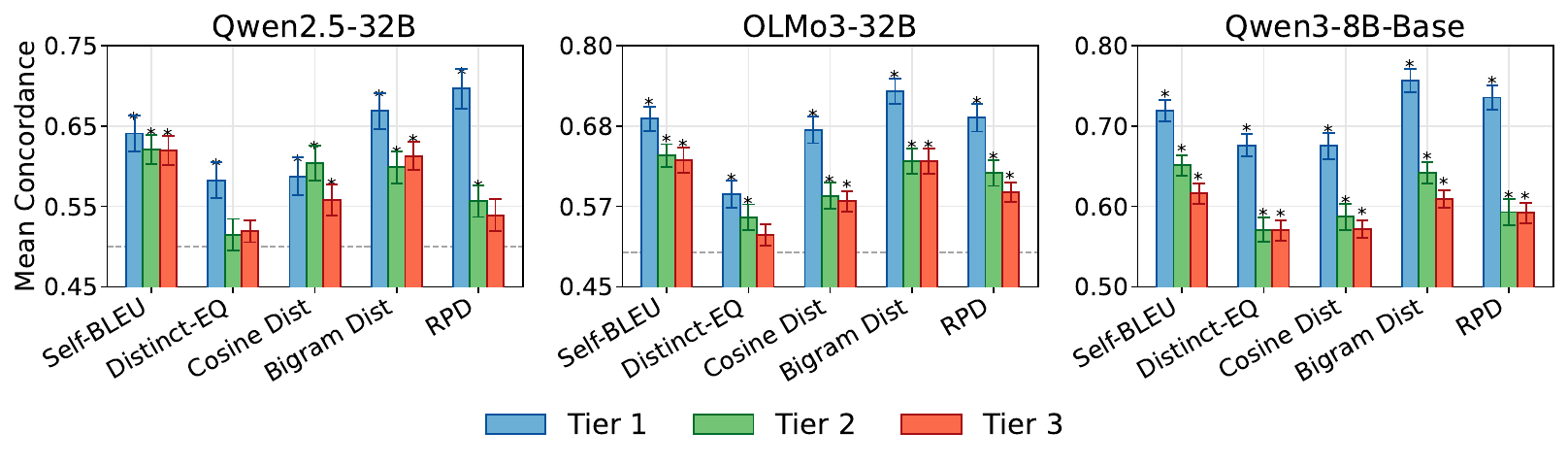}
\caption{\textbf{Set-level concordance} ($C_{\mathrm{set}}$) at subset size $k=4$ across solver models. Bars show mean concordance with 95\% confidence intervals. Asterisks indicate values that are significantly above the random baseline at $p=0.05$.}
\label{fig:ccc_k4_result}
\end{figure*}

\subsection{Existing Measures Fail to Capture Fine-Grained Differences}

Using our evaluation framework, we show that the limitations in 
Section~\ref{sec:fail_model_cov_measures} persist across a broader 
set of diversity measures introduced in Section~\ref{sec:metrics_under_eval}: prior metrics detect whether a solution set uses one approach or many, but lose discriminative power on finer differences within approach-diverse sets.

\paragraph{Concordance analysis}
We evaluate each diversity metric $D$ with a simple ranking question: given two solution sets, how often does $D$ assign the higher score to the more approach-diverse one?
We denote this measure as \textit{concordance}.
For each problem, the LLM judge first clusters correct solutions by approach.
We then sample equal-sized pairs $(S_{\mathrm{high}}, S_{\mathrm{low}})$ with $|S_{\mathrm{high}}|=|S_{\mathrm{low}}|=k$, where $S_{\mathrm{high}}$ covers more distinct approaches than $S_{\mathrm{low}}$. The solutions are distributed as evenly as possible across distinct approaches.
For each question, we repeat the test $B$ times, and count how often the metric gets it right,
\begin{equation}
\label{eq:set_concordance}
C_{\mathrm{set}}(D) = \frac{1}{B} \sum_{b=1}^{B}
\mathds{1}\left[D(S_{\mathrm{high}}^{(b)}) > D(S_{\mathrm{low}}^{(b)})\right],
\end{equation}
\noindent which is then averaged over the problems. For pairwise distance metrics, $D(S)$ denotes the mean pairwise distance within $S$.
Solutions are drawn from three base models: Qwen2.5-32B-Base \citep{qwen2.5}, Qwen3-8B-Base \citep{qwen3technicalreport}, and OLMo3-32B-Base \citep{olmo2026olmo3}. 
We report results with $k=4$ in the main text.

\paragraph{From coarse to fine-grained comparisons}
The easy case is when $S_{\mathrm{low}}$ uses just one approach, 
where solutions look nearly identical, so even surface-level metrics 
can tell them apart from a varied $S_{\mathrm{high}}$.
The more important question is whether a metric can detect a finer 
difference when $S_\mathrm{low}$ itself already induces substantial 
surface diversity.

To separate these cases, we use a tiered evaluation.
At Tier~$n$, $S_{\mathrm{low}}$ contains solutions from $n$ distinct 
approaches, while $S_{\mathrm{high}}$ contains $n+1$ distinct approaches.
Tier~1 then compares one approach against two, acting as a coarse 
contrast.
Tiers~2 and~3 are stricter: they ask whether a metric can detect 
the incremental gain from adding one more approach to an already 
multi-approach set.
These higher tiers test whether a metric truly captures 
approach-level diversity beyond surface form.

\paragraph{Results}
Figure~\ref{fig:ccc_k4_result} shows a consistent pattern across metrics and generator models. Existing metrics perform well in Tier~1, where the comparison is between a single-approach set and a multi-approach set. However, their concordance drops substantially in Tiers 2 and 3, where both sets already contain multiple approaches. Thus, prior metrics can detect coarse diversity contrasts, but struggle to resolve incremental gains in approach-level diversity once both solution sets are already diverse. We experiment with more ablations such as extended math reasoning benchmarks, more embedding models, and different aggregation methods in Appendix~\ref{app:detail_align_eval}. The results consistently shows the same pattern under all settings.


\begin{table}[t]
    \centering
    \scriptsize
    \setlength{\tabcolsep}{2.2pt}
    \setlength{\abovecaptionskip}{3pt}
    \setlength{\belowcaptionskip}{-2pt}
    \renewcommand{\arraystretch}{0.92}
    \resizebox{\columnwidth}{!}{%
    \begin{tabular}{@{}lcccccc@{}}
    \toprule
    & \multicolumn{3}{c}{Qwen2.5-3B}
    & \multicolumn{3}{c}{Qwen2.5-7B} \\
    \cmidrule(lr){2-4} \cmidrule(lr){5-7}
    & \multicolumn{2}{c}{Quality}
    & Diversity
    & \multicolumn{2}{c}{Quality}
    & Diversity \\
    \cmidrule(lr){2-3} \cmidrule(lr){4-4}
    \cmidrule(lr){5-6} \cmidrule(lr){7-7}
    & Pass@1 & Pass@32 & Cov@16
    & Pass@1 & Pass@32 & Cov@16 \\
    \midrule
    Base     & 15.19 & 60.30 & 2.302 & 19.10 & 66.52 & 1.956 \\
    \arrayrulecolor{gray!65}\midrule[0.4pt]\arrayrulecolor{black}
    GRPO     & 27.81 & 64.44 & 1.786 & 37.08 & 71.70 & 1.761 \\
    \arrayrulecolor{gray!65}\midrule[0.4pt]\arrayrulecolor{black}
    DQO      & 26.48 & 63.85 & 1.853 & 31.86 & 66.20 & 1.508 \\
    DIVER-TD & 28.22 & 65.93 & 1.519 & 38.36 & 72.19 & 1.524 \\
    DIVER-ED & 29.31 & 66.67 & 1.884 & 39.42 & 71.11 & 1.573 \\
    \bottomrule
    \end{tabular}%
    }
    \caption{Quality and diversity comparison across different diversity-aware RL algorithms.}
    \label{tab:diversity_rl}
\end{table}

\begin{figure*}[t]
    \centering

    \begin{subfigure}[t]{0.48\textwidth}
        \centering
        \includegraphics[width=\linewidth]{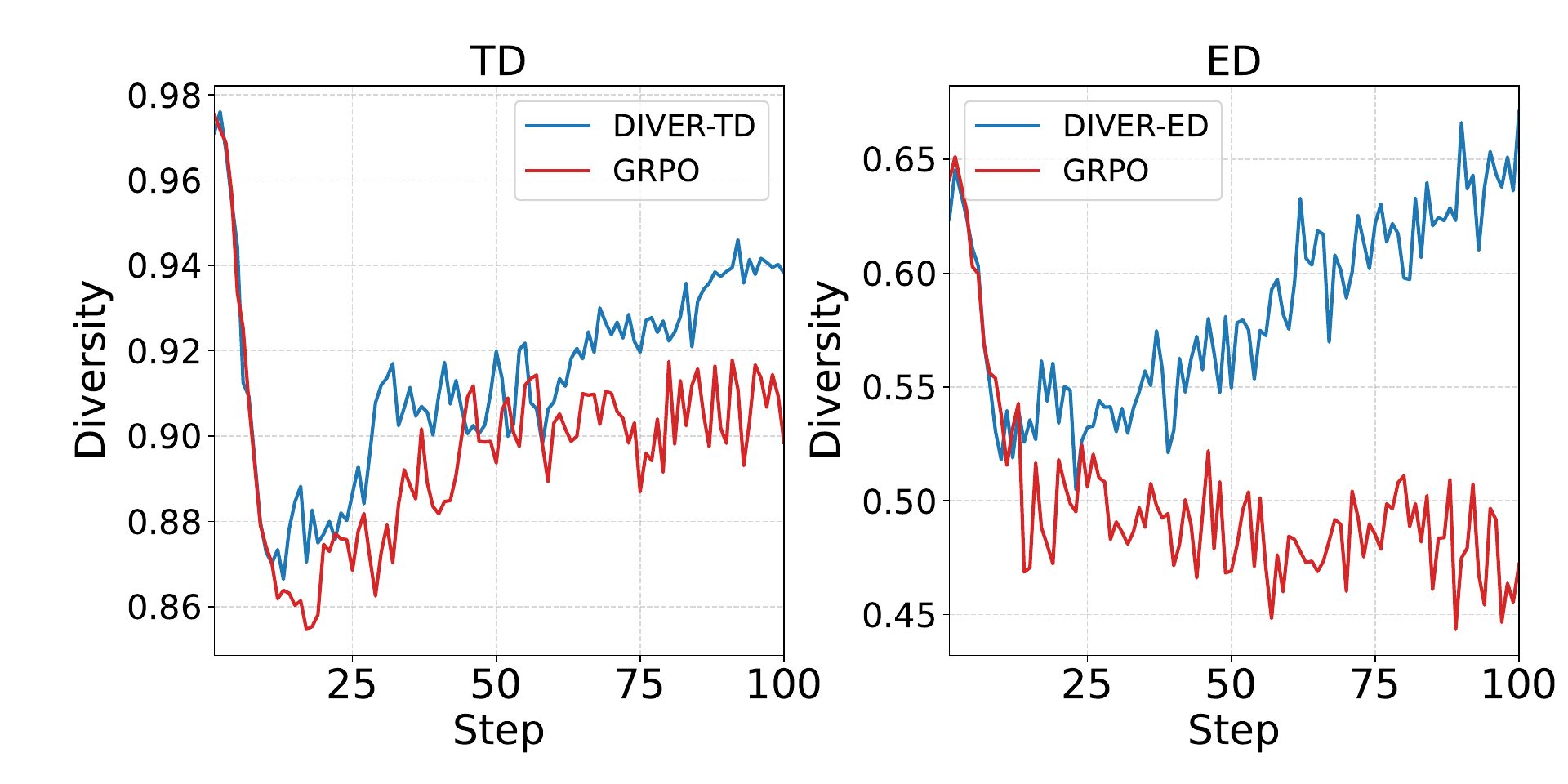}
        \caption{Training diversity of DIVER.}
        \label{fig:diver-training-diversity}
    \end{subfigure}
    \hfill
    {%
    \begin{subfigure}[t]{0.24\textwidth}
        \centering
        \includegraphics[width=\linewidth]{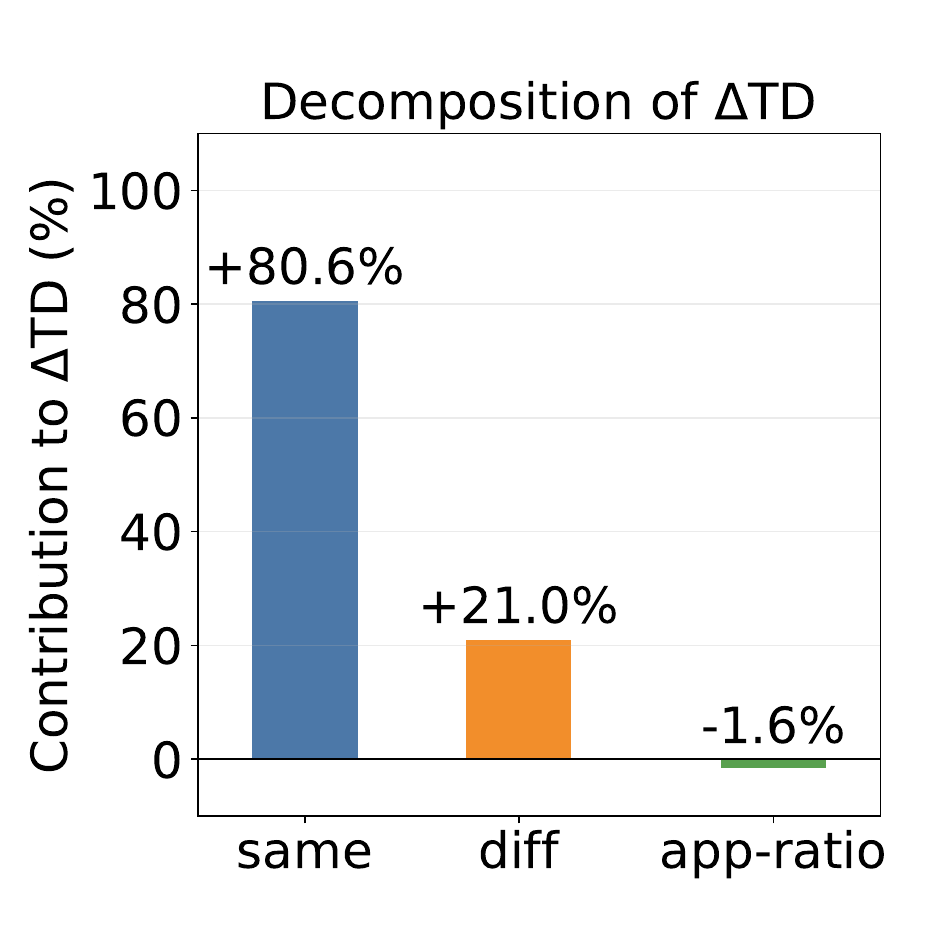}
        \caption{Textual diversity gain}
        \label{fig:change-on-delta-bleu}
    \end{subfigure}
    }
    \hfill
    {%
    \begin{subfigure}[t]{0.24\textwidth}
        \centering
        \includegraphics[width=\linewidth]{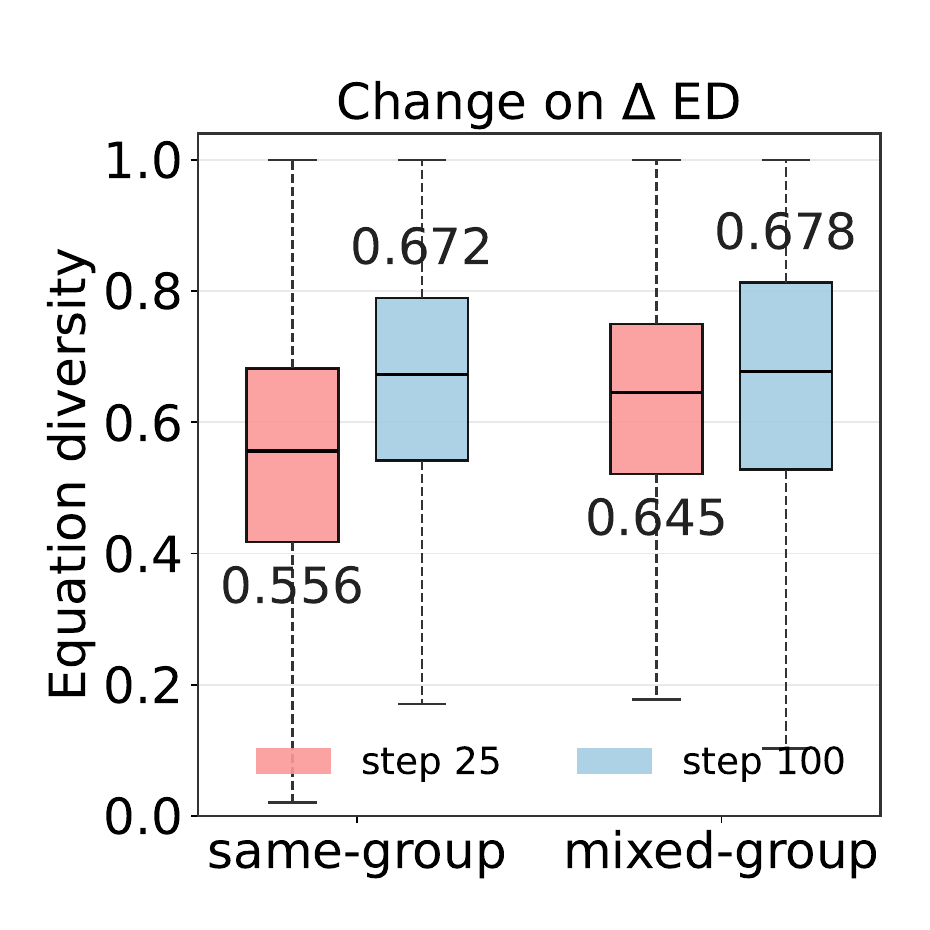}
        \caption{Equation diversity gain}
        \label{fig:change-on-delta-equ}
    \end{subfigure}
    }

    \caption{\textbf{Analysis of DIVER.}
DIVER preserves its target TD and ED metrics, but the recovered diversity mostly reflects within-approach variation rather than broader approach coverage.
See Appendix~\ref{app:diver_analysis_detail} for details.}
    \label{fig:combined-three-figures}
\end{figure*}

\subsection{What Diversity Does Diversity-Aware RLVR Preserve?}
\label{sec:diversity_rl}

Diversity-aware RLVR methods aim to preserve generation diversity, but 
typically optimize surface-level measures. 
Given the gap between these measures 
and approach-level diversity, we ask whether such methods broaden the 
set of approaches or merely produce surface variants of the same ones. 
We study two recent baselines: DQO 
\citep{chen2026dqo}, which rewards embedding diversity, and DIVER 
\citep{hu2026diver}, which combines textual diversity (TD) via pairwise 
BLEU with equation diversity (ED) via distinct equations. 
Appendix~\ref{app:diver_analysis_detail} provides full descriptions.


\paragraph{Measuring approach coverage}
We measure approach-level diversity of a policy $\pi$ with \textit{expected coverage} $\mathrm{cov}@N$: the expected number of distinct approach clusters observed when sampling $N$ correct solutions from $\pi$.
This captures not only the number of covered approaches, but also how evenly the policy samples across them.
For a problem $x$, we define this over sets $S_x$ of $N$ correct samples:
\begin{equation}
\mathrm{cov}_x(N, \pi) = \mathbb{E}_{S_x} \left[ | \mathcal{J}(x, S_x) | \right], \quad |S_x| = N,
\end{equation}
where $\mathcal{J}(x, S_x)$ is the clustering produced by the judge from Section~\ref{sec:llm_diversity_judge}. We average this value across validation problems to obtain a policy-level coverage score $\mathrm{cov}@N$. Estimation details are deferred to Appendix~\ref{app:coverage}.

\paragraph{Experimental setup}
We train Qwen2.5-3B/7B-Base models on the MATH training set for 100 steps.
For each trained policy, we measure quality by accuracy on OlympiadBench \citep{olympiadbench}.
For diversity, we evaluate approach coverage on 150 problems from the approach-feasible problem set in Section~\ref{sec:multi_approach_problems}, retaining only those for which every Qwen2.5-3B checkpoint, including the base model, produces at least 16 correct solutions.
This restriction ensures that coverage is estimated from enough correct samples for a reliable comparison.

\paragraph{Results}
Table~\ref{tab:diversity_rl} shows that approach-level diversity declines after RLVR across all settings, indicating that RLVR’s diversity decline also appears at the approach level.
Moreover, preserving the optimized proxy does not imply preserving approach-level diversity: DIVER maintains its textual or equation-level diversity signals during training (Figure~\ref{fig:diver-training-diversity}), but still loses approach coverage.

To understand what these targeted gains represent, we decompose DIVER’s target-metric improvements between steps 25 and 100, where the optimized metrics increase. The decomposition shows that the gains come mainly from making solutions within the same approach more varied, rather than from producing more diverse approaches. In Figure~\ref{fig:change-on-delta-bleu}, 80.6\% of the textual-diversity gain comes from greater variation among same-approach pairs. Figure~\ref{fig:change-on-delta-equ} shows a similar pattern for equations: same-approach equation diversity rises sharply, while cross-approach equation diversity remains relatively consistent.

These results reveal an implication: optimizing surface-level diversity can make generations look more diverse without widening the set of strategies the policy uses. Instead, it produces varied realizations of a narrower set of reasoning strategies. Thus, the surface–approach diversity gap also reflects a limitation of the surface-level objective rather than merely an evaluation artifact.
Further details on the analysis and additional experiments, including variance across training seeds and longer training runs, are provided in Appendix~\ref{app:details_diversity_RLVR}.

\section{The Utility and Limits of Optimizing Approach-Level Diversity}
\label{sec:approach_utility}


Having shown that conventional metrics misalign with approach-level 
diversity, we now ask two questions: does approach-level diversity 
help downstream performance, and can we induce it directly? 
We address the first through approach-controlled test-time scaling, 
and the second through reward optimization with an LLM judge.

\begin{figure}[t]
  \centering
  \includegraphics[width=\linewidth]{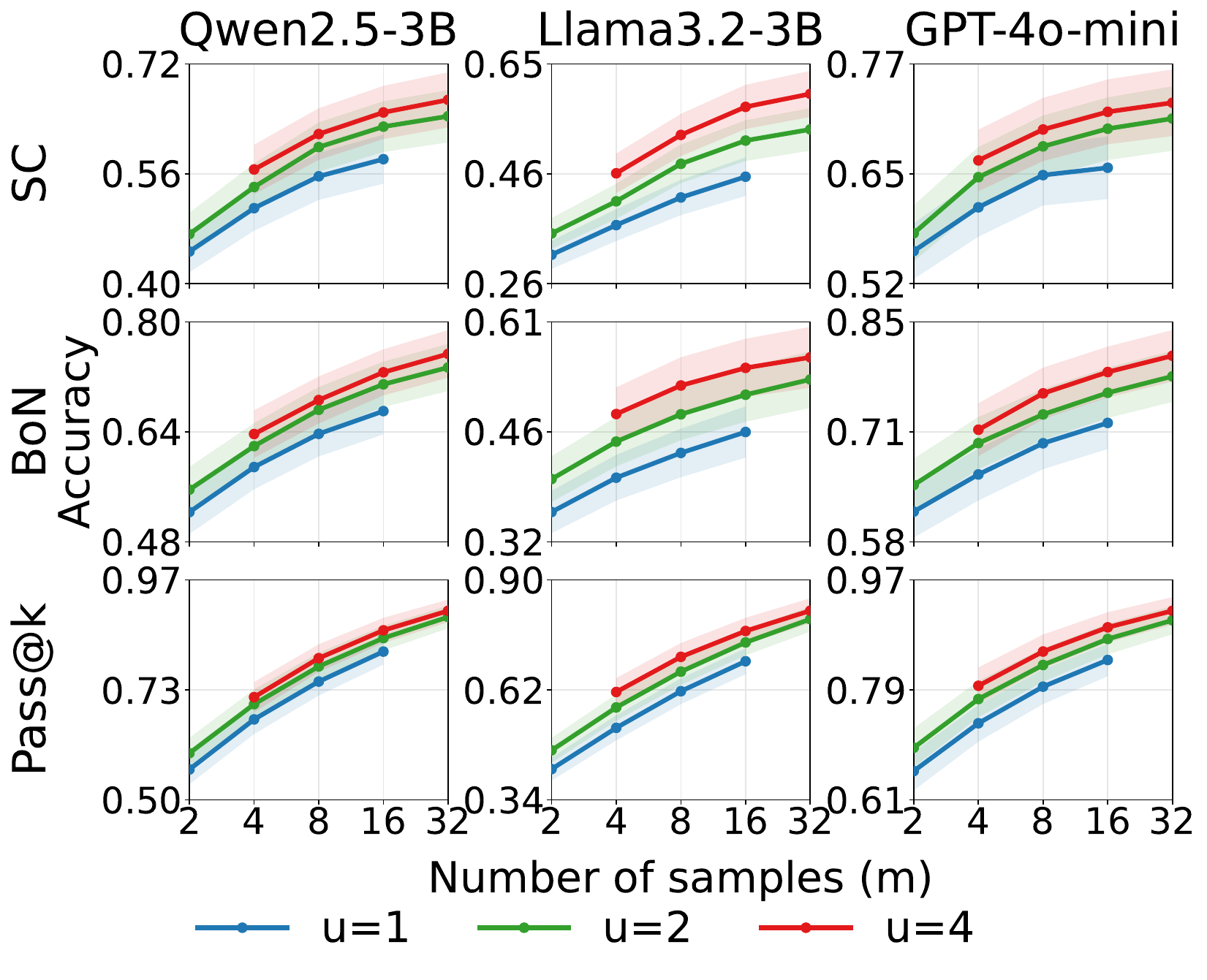}
  \caption{\textbf{Approach-controlled test-time scaling} with varying numbers of unique approaches. 95\% confidence intervals are computed across problems.}
  \label{fig:approach-count}
\end{figure}

\subsection{Approach diversity for Test-Time Scaling}

We first test whether approach-level diversity improves test-time scaling.
Given equally sized candidate sets that differ only in the number of covered approaches, we compare three representative scaling methods: self-consistency \citep{selfconsistency}, best-of-$N$ \citep{lightman2023letsverifystepstep}, and pass@k, using a setup inspired by \citet{handa2026guidedsample}.

\paragraph{Approach-controlled subsets}
For this experiment, we use the approach-feasible set from Section~\ref{sec:multi_approach_problems}, where \(x\) is associated with valid and distinct approach plans \(C=\{c_1,\dots,c_K\}\).
For each plan \(c_k\), we prompt a solver model \(p_\theta\) to generate \(R\) independent solutions,
\[
  S = \bigcup_{k=1}^{K} S_k, \quad S_k = \left\{ s_{kj} \sim p_\theta(\cdot \mid x, c_k) \right\}_{j=1}^{R}.
\]
We then form subsets \(\tilde{S}_{m,u} \subset S\), where \(m\) is the total number of solutions and \(u\) is the number of covered approaches.
For fixed \(m\), we vary \(u\) by sampling \(u\) approach plans and drawing \(m/u\) solutions from each.
By construction, the expected single-solution quality is the same among subsets, so higher performance after test-time scaling directly indicates higher gains. 

\paragraph{Results}
Across three model families---Qwen2.5-3B-Instruct \citep{qwen2.5}, Llama3.2-3B-Instruct \citep{llama3modelcard}, and GPT-4o-mini \citep{gpt-4o}---Figure~\ref{fig:approach-count} shows a consistent trend: covering more approaches leads to better test-time scaling, achieving higher performance for a given sampling budget, suggesting that approach-level diversity provides practical utility at inference time.
However, this utility is not applicable to arbitrary problems. Appendix~\ref{app:approach_seeking_sft} shows that fine-tuning a model to seek multiple approaches improves accuracy primarily on problems where the alternative approaches are \emph{reachable}, i.e., the model can solve the problem by following each of approach, implying that approach-level diversity pays off only when the model can execute those approaches.


Exact numerical values and additional results with non-oracle, self-generated labels are provided in Appendix~\ref{app:test_time_scaling_details}.

\begin{table}[t]
    \centering
    \small
    \setlength{\tabcolsep}{3.0pt}
    \setlength{\abovecaptionskip}{3pt}
    \setlength{\belowcaptionskip}{-2pt}
    \renewcommand{\arraystretch}{1.15}
    \begin{tabular}{@{}lcc>{\centering\arraybackslash}p{1.65cm}>{\centering\arraybackslash}p{1.65cm}@{}}
    \toprule
    & \multicolumn{2}{c}{Quality} & \multicolumn{2}{c}{Diversity (Cov@16)} \\
    \cmidrule(lr){2-3} \cmidrule(lr){4-5}
    & Pass@1 & Pass@32 & \makecell{GPT\\\scriptsize(held-out)} & \makecell{Qwen\\\scriptsize(in-loop)} \\
    \midrule
    Base              & 15.19 & 60.30 & 2.302 & 2.031 \\
    \arrayrulecolor{gray!50}\midrule\arrayrulecolor{black}
    GRPO              & 27.81 & 64.44 & 1.786 & 1.702 \\
    \arrayrulecolor{gray!50}\midrule\arrayrulecolor{black}
    \multicolumn{5}{@{}l}{\textcolor{gray}{\scriptsize{Judge reward}}} \\
    Sqrt        & 29.71 & 65.04 & 1.426 & 1.657 \\
    Sqrt+prior  & 29.43 & 64.30 & 1.499 & 1.786 \\
    Linear      & 29.53 & 66.22 & 1.492 & 1.708 \\
    \bottomrule
    \end{tabular}
    \caption{Results of training with LLM-judge-based diversity rewards under the DIVER framework. Quality is measured by accuracy on OlympiadBench. Diversity is measured by Cov@16 with two judges: Qwen, the \textbf{in-loop judge} used to compute the training reward, and GPT, a \textbf{held-out judge} not used during training.}
    \label{tab:diver_judge}
\end{table}

\subsection{Optimizing an LLM Judge Reward Amplifies Judge Bias}
\label{sec:llm_judge_rlvr}

\begin{figure}[t]
  \centering
  \includegraphics[width=\linewidth]{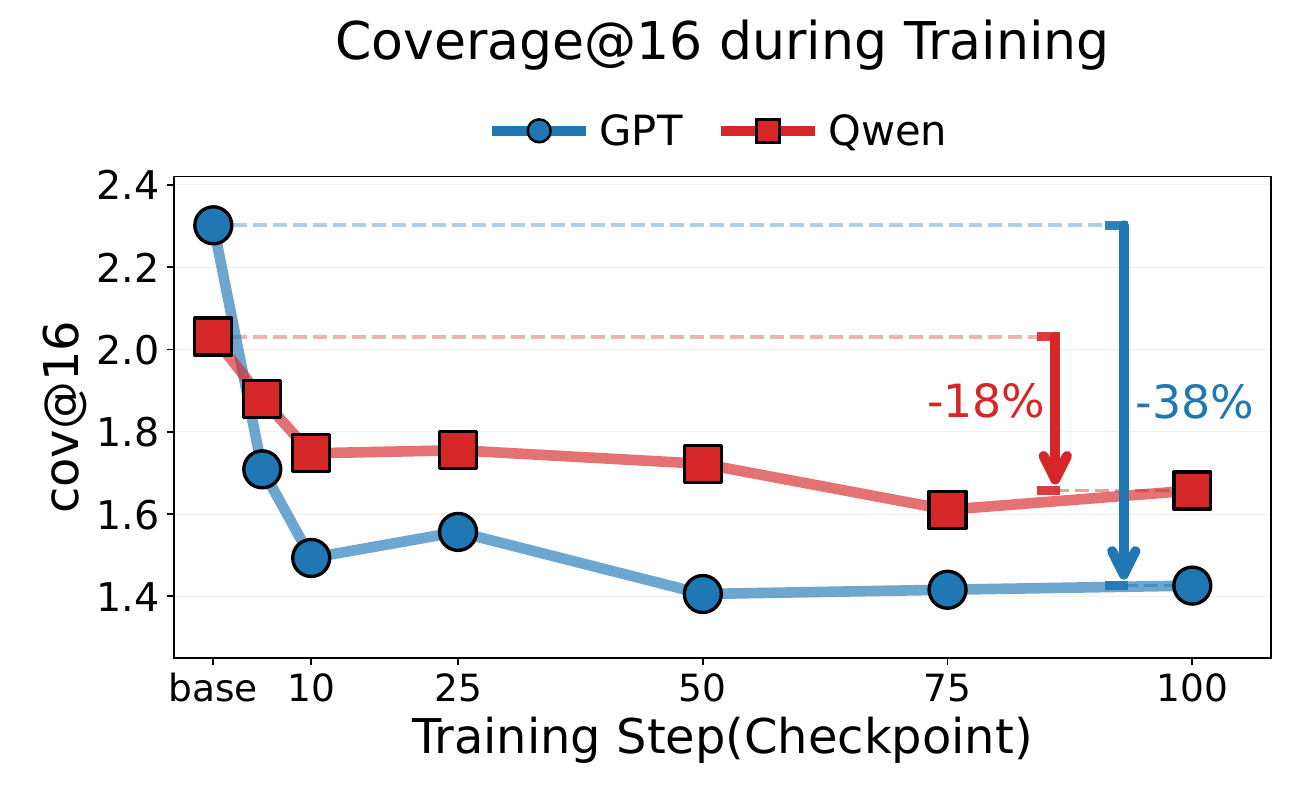}
  \caption{
  Approach-level diversity (Cov@16) on the evaluation set across training checkpoints, as measured by the \textbf{in-loop Qwen judge} (red) and the \textbf{held-out GPT judge} (blue).
  }
  \label{fig:judge_mismatch_coverage_main}
\end{figure}

We next ask whether approach-level diversity can be directly encouraged 
during training through an LLM-judge-based reward. 
Section~\ref{sec:diversity_rl} showed that surface-level rewards 
maintain their target proxies but that approach-level diversity declines. A natural question is to reward approach-level diversity 
directly, using an LLM-judge-based reward. 
We instantiate this with a 35B open-source judge, but find that direct 
optimization fails in a different way: the policy learns judge-specific 
preferences rather than broadening its set of approaches.

\paragraph{Setup}
We use the judge from Section~\ref{sec:llm_diversity_judge}, replacing the backbone with Qwen3.5-35B-A3B~\cite{qwen35blog}, as using a proprietary judge during training is costly.
Given a problem and the correct rollouts, the judge clusters them into approaches following Equation~\ref{eq:clustering_equation}.
Each correct rollout \(s_i \in C_j\) then receives a diversity reward \(r_{\mathrm{div}}\) determined by the size of its cluster, \(n = |C_j|\).
We consider three formulations of \(r_{\mathrm{div}}\):
\begin{enumerate}
    \item \textbf{Sqrt}: \(r_{\mathrm{div}} = 1/\sqrt{n}\), where total per problem is maximized under a balanced distribution across approaches, aligning with our expected-coverage metric.
    \item \textbf{Sqrt+prior}: \(r_{\mathrm{div}} = \sqrt{(1+\alpha)/(n+\alpha)}\), a smoothed variant of Sqrt with prior \(\alpha\) that accounts for the limited number of rollouts per problem.
    \item \textbf{Linear}: \(r_{\mathrm{div}} = 1/n\), a variant with sharper decay, also explored in prior work~\citep{li2025jointly}.
\end{enumerate}
We plug each reward into DIVER, keep the rest of the training setup from Section~\ref{sec:diversity_rl}, and train Qwen2.5-3B on the MATH training set.

\paragraph{Results}
Despite optimizing an explicit approach-diversity reward, judge-reward training fails to improve approach-level diversity.
Under our human-calibrated GPT judge, all three variants yield \emph{lower diversity than quality-only GRPO} (Table~\ref{tab:diver_judge}).
Moreover, the two judges disagree in a revealing manner: for policies never trained against the Qwen judge (Base and GRPO), GPT reports higher diversity, whereas for every judge-reward variant the Qwen judge assigns a higher diversity score. 

Figure~\ref{fig:judge_mismatch_coverage_main} shows how this reversal emerges during training with the Sqrt reward.
Before training, GPT rates rollouts as more diverse than Qwen; as training proceeds, diversity under the in-loop Qwen judge degrades only modestly (18\%), while diversity under the held-out GPT judge collapses sharply (38\%), flipping the ordering within the first few steps.
This suggests that the policy is learning to produce rollouts that are distinct only to the in-loop judge, while an independent judge sees the underlying strategies collapsing. 

Naively plugging an LLM judge into the reward therefore cannot preserve approach-level diversity, leaving an open challenge: designing diversity signals that capture genuine strategy-level differences and remain robust to when optimized against.

\section{Discussions}

\paragraph{Future Research Directions} 
Our results suggest two directions for future work. 
First, useful diversity is likely to become increasingly \textit{domain-specific} as LLMs are applied to more complex tasks. While mathematical reasoning offers a controlled setting, domains such as scientific discovery, program synthesis, and open-ended planning may demand different notions of strategic variation, motivating domain-aware definitions and evaluations of approach diversity. 
Second, approach-level diversity can be further studied as a \textit{training objective}. 
Our evaluation framework makes it possible to identify and analyze approach-level variation at scale, opening a path toward diversity-aware optimization. 
The key challenge is designing objectives that preserve the human-relevant structure of approach-level diversity while remaining robust under direct optimization—a step toward training methods that encourage genuinely distinct reasoning rather than surface-level variation.

\paragraph{Conclusion}
We introduce approach-level diversity as a distinct axis for analyzing mathematical reasoning in LLMs. Our results show that existing metrics and diversity-aware training methods often conflate surface variation with genuinely distinct solution strategies. 
By making this gap explicit, we believe our analysis points toward training and evaluation that better reflect human-aligned notions of diversity in mathematical reasoning.



\section*{Limitations}
\paragraph{Focus on Mathematical Reasoning}
The experiments and the analysis of this paper is focused to closed-ended mathematical reasoning at the small to medium sized models model. 
We focus on such because the distinction between surface form and approach-level structure is relatively clear, and because recent diversity-aware RLVR methods are largely studied on math benchmarks. 
Moreover, while our judge-reward RLVR experiments reveal a concerning potential failure mode, they are limited to a single model scale, and experiments across larger model sizes are needed to determine whether this phenomenon generalizes.
A natural next step is to extend this analysis to domains where higher-level strategic diversity may provide more direct benefits, such as program synthesis, scientific discovery, or open-ended planning.

\paragraph{Non-exhaustive Coverage}
We evaluate representative diversity metrics and recent diversity-aware RLVR methods, but do not exhaustively cover all possible metrics, training objectives, or decoding strategies. Our results should therefore be read as evidence of a mismatch in commonly used diversity proxies, rather than as an exhaustive survey of diversity methods.

\paragraph{No General Mitigation Method}
Our work identifies and analyzes the mismatch between diversity proxies and approach diversity, but does not provide a general method for preserving or inducing approach diversity during training. Future work may require both more faithful reward signals and training algorithms that can optimize them without proxy exploitation.

\newpage

\section*{Acknowledgments}
This work was mainly supported by the Institute of Information \& communications Technology Planning \& Evaluation (IITP) grant funded by the Korea government(MSIT) (No.RS-2022-II220184, Development and Study of AI Technologies to Inexpensively Conform to Evolving Policy on Ethics), 
IITP (Institute of Information \& Communications Technology Planning \& Evaluation)-ITRC (Information Technology Research Center) grant funded by the Korea government (Ministry of Science and ICT) (IITP-2026-RS-2024-00437633). 
K. Jung is with ASRI, Seoul National University, Korea.



\bibliography{custom}

@misc{handa2026guidedsample,
      title={GuidedSampling: Steering LLMs Towards Diverse Candidate Solutions at Inference-Time}, 
      author={Divij Handa and Mihir Parmar and Aswin RRV and Md Nayem Uddin and Hamid Palangi and Chitta Baral},
      year={2026},
      eprint={2510.03777},
      archivePrefix={arXiv},
      primaryClass={cs.AI},
      url={https://arxiv.org/abs/2510.03777}, 
}

@misc{ju2026reasoningpathdivergencenew,
      title={Reasoning Path Divergence: A New Metric and Curation Strategy to Unlock LLM Diverse Thinking}, 
      author={Feng Ju and Zeyu Qin and Rui Min and Zhitao He and Lingpeng Kong and Yi R. Fung},
      year={2026},
      eprint={2510.26122},
      archivePrefix={arXiv},
      primaryClass={cs.CL},
      url={https://arxiv.org/abs/2510.26122}, 
}

@inproceedings{wuprogress,
  title={Progress or Regress? Self-Improvement Reversal in Post-training},
  author={Wu, Ting and Li, Xuefeng and Liu, Pengfei},
  booktitle={The Thirteenth International Conference on Learning Representations},
  year={2024}
}

@inproceedings{zhu2018texygen,
  title={Texygen: A benchmarking platform for text generation models},
  author={Zhu, Yaoming and Lu, Sidi and Zheng, Lei and Guo, Jiaxian and Zhang, Weinan and Wang, Jun and Yu, Yong},
  booktitle={The 41st international ACM SIGIR conference on research \& development in information retrieval},
  pages={1097--1100},
  year={2018}
}

@inproceedings{kondrak2005n,
  title={N-gram similarity and distance},
  author={Kondrak, Grzegorz},
  booktitle={International symposium on string processing and information retrieval},
  pages={115--126},
  year={2005},
  organization={Springer}
}

@inproceedings{wu2025invisible,
  title={The Invisible Leash: Why RLVR May Not Escape Its Origin},
  author={Wu, Fang and Choi, Yejin},
  booktitle={2nd AI for Math Workshop@ ICML 2025},
  year={2025}
}

@article{yuedoes,
  title={Does Reinforcement Learning Really Incentivize Reasoning Capacity in LLMs Beyond the Base Model?},
  author={Yue, Yang and Chen, Zhiqi and Lu, Rui and Zhao, Andrew and Wang, Zhaokai and Song, Shiji and Huang, Gao},
  year={2025}
}

@article{cuientropy,
  title={The Entropy Mechanism of Reinforcement Learning for Reasoning Language Models},
  author={Cui, Ganqu and Zhang, Yuchen and Chen, Jiacheng and Yuan, Lifan and Wang, Zhi and Zuo, Yuxin and Li, Haozhan and Fan, Yuchen and Chen, Huayu and Chen, Weize and others},
  year={2025}
}

@inproceedings{hu2026diver,
  title={Diversity-incentivized exploration for versatile reasoning},
  author={Hu, Zican and Zhang, Shilin and Li, Yafu and Yan, Jianhao and Hu, Xuyang and Cui, Leyang and Qu, Xiaoye and Chen, Chunlin and Cheng, Yu and Wang, Zhi},
  booktitle={The Fourteenth International Conference on Learning Representations},
  year={2026}
}

@inproceedings{chen2026dqo,
  title={Post-training Large Language Models for Diverse High-Quality Responses},
  author={Chen, Yilei and Chakraborty, Souradip and Wolf, Lorenz and Paschalidis, Yannis and Pacchiano, Aldo},
  booktitle={The Fourteenth International Conference on Learning Representations},
  year={2026}
}

@article{li2025jointly,
  title={Jointly reinforcing diversity and quality in language model generations},
  author={Li, Tianjian and Zhang, Yiming and Yu, Ping and Saha, Swarnadeep and Khashabi, Daniel and Weston, Jason and Lanchantin, Jack and Wang, Tianlu},
  journal={arXiv preprint arXiv:2509.02534},
  year={2025}
}

@misc{jiang2025artificialhivemindopenendedhomogeneity,
      title={Artificial Hivemind: The Open-Ended Homogeneity of Language Models (and Beyond)}, 
      author={Liwei Jiang and Yuanjun Chai and Margaret Li and Mickel Liu and Raymond Fok and Nouha Dziri and Yulia Tsvetkov and Maarten Sap and Alon Albalak and Yejin Choi},
      year={2025},
      eprint={2510.22954},
      archivePrefix={arXiv},
      primaryClass={cs.CL},
      url={https://arxiv.org/abs/2510.22954}, 
}

@misc{puri2026reachingmoderldistributional,
      title={Reaching Beyond the Mode: RL for Distributional Reasoning in Language Models}, 
      author={Isha Puri and Mehul Damani and Idan Shenfeld and Marzyeh Ghassemi and Jacob Andreas and Yoon Kim},
      year={2026},
      eprint={2603.24844},
      archivePrefix={arXiv},
      primaryClass={cs.LG},
      url={https://arxiv.org/abs/2603.24844}, 
}

@inproceedings{olympiadbench,
  title={Olympiadbench: A challenging benchmark for promoting agi with olympiad-level bilingual multimodal scientific problems},
  author={He, Chaoqun and Luo, Renjie and Bai, Yuzhuo and Hu, Shengding and Thai, Zhen and Shen, Junhao and Hu, Jinyi and Han, Xu and Huang, Yujie and Zhang, Yuxiang and others},
  booktitle={Proceedings of the 62nd Annual Meeting of the Association for Computational Linguistics (Volume 1: Long Papers)},
  pages={3828--3850},
  year={2024}
}

@misc{aime,
  author       = {{Mathematical Association of America}},
  title        = {{A}merican {I}nvitational {M}athematics {E}xamination ({AIME})},
  year         = {2024},
  howpublished = {\url{https://artofproblemsolving.com/wiki/index.php/AIME_Problems_and_Solutions}}
}

@article{hendrycksmath,
  title={Measuring mathematical problem solving with the math dataset},
  author={Hendrycks, Dan and Burns, Collin and Kadavath, Saurav and Arora, Akul and Basart, Steven and Tang, Eric and Song, Dawn and Steinhardt, Jacob},
  journal={arXiv preprint arXiv:2103.03874},
  year={2021}
}

@inproceedings{selfconsistency,
  title={Self-Consistency Improves Chain of Thought Reasoning in Language Models},
  author={Wang, Xuezhi and Wei, Jason and Schuurmans, Dale and Le, Quoc V and Chi, Ed H and Narang, Sharan and Chowdhery, Aakanksha and Zhou, Denny},
  booktitle={The Eleventh International Conference on Learning Representations},
  year={2023}
}

@inproceedings{zhang2025noveltybench,
  title={NoveltyBench: Evaluating Language Models for Humanlike Diversity},
  author={Zhang, Yiming and Diddee, Harshita and Holm, Susan and Liu, Hanchen and Liu, Xinyue and Samuel, Vinay and Wang, Barry and Ippolito, Daphne},
  booktitle={Conference on Language Modeling (COLM)},
  year={2025}
}

@inproceedings{he2025rewarding,
  title={Rewarding the unlikely: Lifting grpo beyond distribution sharpening},
  author={He, Andre Wang and Fried, Daniel and Welleck, Sean},
  booktitle={Proceedings of the 2025 Conference on Empirical Methods in Natural Language Processing},
  pages={25559--25571},
  year={2025}
}

@article{lanchantin2025diverse,
  title={Diverse preference optimization},
  author={Lanchantin, Jack and Chen, Angelica and Dhuliawala, Shehzaad and Yu, Ping and Weston, Jason and Sukhbaatar, Sainbayar and Kulikov, Ilia},
  journal={arXiv preprint arXiv:2501.18101},
  year={2025}
}

@article{chen2025dra,
  title={Dra-grpo: Exploring diversity-aware reward adjustment for r1-zero-like training of large language models},
  author={Chen, Xiwen and Zhu, Wenhui and Qiu, Peijie and Dong, Xuanzhao and Wang, Hao and Wu, Haiyu and Li, Huayu and Sotiras, Aristeidis and Wang, Yalin and Razi, Abolfazl},
  journal={arXiv preprint arXiv:2505.09655},
  year={2025}
}

@inproceedings{kirkunderstanding,
  title={Understanding the Effects of RLHF on LLM Generalisation and Diversity},
  author={Kirk, Robert and Mediratta, Ishita and Nalmpantis, Christoforos and Luketina, Jelena and Hambro, Eric and Grefenstette, Edward and Raileanu, Roberta},
  booktitle={The Twelfth International Conference on Learning Representations},
  year={2024},
}

@article{shumailov2024ai,
  title={{AI} models collapse when trained on recursively generated data},
  author={Shumailov, Ilia and Shumaylov, Zakhar and Zhao, Yiren and Papernot, Nicolas and Anderson, Ross and Gal, Yarin},
  journal={Nature},
  volume={631},
  number={8022},
  pages={755--759},
  year={2024},
  doi={10.1038/s41586-024-07566-y}
}

@inproceedings{wu2024autogen,
  title={Autogen: Enabling next-gen LLM applications via multi-agent conversations},
  author={Wu, Qingyun and Bansal, Gagan and Zhang, Jieyu and Wu, Yiran and Li, Beibin and Zhu, Erkang and Jiang, Li and Zhang, Xiaoyun and Zhang, Shaokun and Liu, Jiale and others},
  booktitle={First Conference on language modeling},
  year={2024}
}

@article{shao2024deepseekmath,
  title={Deepseekmath: Pushing the limits of mathematical reasoning in open language models},
  author={Shao, Zhihong and Wang, Peiyi and Zhu, Qihao and Xu, Runxin and Song, Junxiao and Bi, Xiao and Zhang, Haowei and Zhang, Mingchuan and Li, YK and others},
  journal={arXiv preprint arXiv:2402.03300},
  year={2024}
}

@misc{qwen2.5,
    title = {Qwen2.5: A Party of Foundation Models},
    url = {https://qwenlm.github.io/blog/qwen2.5/},
    author = {Qwen Team},
    month = {September},
    year = {2024}
}

@misc{qwen3technicalreport,
      title={Qwen3 Technical Report}, 
      author={Qwen Team},
      year={2025},
      eprint={2505.09388},
      archivePrefix={arXiv},
      primaryClass={cs.CL},
      url={https://arxiv.org/abs/2505.09388}, 
}

@misc{olmo2026olmo3,
      title={Olmo 3}, 
      author={Team Olmo and : and Allyson Ettinger and Amanda Bertsch and Bailey Kuehl and David Graham and David Heineman and Dirk Groeneveld and Faeze Brahman and Finbarr Timbers and Hamish Ivison and Jacob Morrison and Jake Poznanski and Kyle Lo and Luca Soldaini and Matt Jordan and Mayee Chen and Michael Noukhovitch and Nathan Lambert and Pete Walsh and Pradeep Dasigi and Robert Berry and Saumya Malik and Saurabh Shah and Scott Geng and Shane Arora and Shashank Gupta and Taira Anderson and Teng Xiao and Tyler Murray and Tyler Romero and Victoria Graf and Akari Asai and Akshita Bhagia and Alexander Wettig and Alisa Liu and Aman Rangapur and Chloe Anastasiades and Costa Huang and Dustin Schwenk and Harsh Trivedi and Ian Magnusson and Jaron Lochner and Jiacheng Liu and Lester James V. Miranda and Maarten Sap and Malia Morgan and Michael Schmitz and Michal Guerquin and Michael Wilson and Regan Huff and Ronan Le Bras and Rui Xin and Rulin Shao and Sam Skjonsberg and Shannon Zejiang Shen and Shuyue Stella Li and Tucker Wilde and Valentina Pyatkin and Will Merrill and Yapei Chang and Yuling Gu and Zhiyuan Zeng and Ashish Sabharwal and Luke Zettlemoyer and Pang Wei Koh and Ali Farhadi and Noah A. Smith and Hannaneh Hajishirzi},
      year={2026},
      eprint={2512.13961},
      archivePrefix={arXiv},
      primaryClass={cs.CL},
      url={https://arxiv.org/abs/2512.13961}, 
}

@misc{schulman2017ppo,
      title={Proximal Policy Optimization Algorithms}, 
      author={John Schulman and Filip Wolski and Prafulla Dhariwal and Alec Radford and Oleg Klimov},
      year={2017},
      eprint={1707.06347},
      archivePrefix={arXiv},
      primaryClass={cs.LG},
      url={https://arxiv.org/abs/1707.06347}, 
}

@misc{qwen35blog,
    title = {Qwen3.5: Accelerating Productivity with Native Multimodal Agents},
    url = {https://qwen.ai/blog?id=qwen3.5},
    author = {Qwen Team},
    month = {February},
    year = {2026}
}

@inproceedings{zhu2026flowrl,
  title={Flow{RL}: Matching Reward Distributions for {LLM} Reasoning},
  author={Zhu, Xuekai and Cheng, Daixuan and Zhang, Dinghuai and Li, Hengli and Zhang, Kaiyan and Jiang, Che and Sun, Youbang and Hua, Ermo and Zuo, Yuxin and Lv, Xingtai and others},
  booktitle={The Fourteenth International Conference on Learning Representations},
  year={2026},
  url={https://openreview.net/forum?id=lObnTKbm9U}
}

@misc{qwen3embedding,
      title={Qwen3 Embedding: Advancing Text Embedding and Reranking Through Foundation Models}, 
      author={Yanzhao Zhang and Mingxin Li and Dingkun Long and Xin Zhang and Huan Lin and Baosong Yang and Pengjun Xie and An Yang and Dayiheng Liu and Junyang Lin and Fei Huang and Jingren Zhou},
      year={2025},
      eprint={2506.05176},
      archivePrefix={arXiv},
      primaryClass={cs.CL},
      url={https://arxiv.org/abs/2506.05176}, 
}

@misc{lightman2023letsverifystepstep,
      title={Let's Verify Step by Step}, 
      author={Hunter Lightman and Vineet Kosaraju and Yura Burda and Harri Edwards and Bowen Baker and Teddy Lee and Jan Leike and John Schulman and Ilya Sutskever and Karl Cobbe},
      year={2023},
      eprint={2305.20050},
      archivePrefix={arXiv},
      primaryClass={cs.LG},
      url={https://arxiv.org/abs/2305.20050}, 
}

@inproceedings{creativewriting,
    title = "A Confederacy of Models: a Comprehensive Evaluation of {LLM}s on Creative Writing",
    author = "G{\'o}mez-Rodr{\'i}guez, Carlos  and
      Williams, Paul",
    editor = "Bouamor, Houda  and
      Pino, Juan  and
      Bali, Kalika",
    booktitle = "Findings of the Association for Computational Linguistics: EMNLP 2023",
    month = dec,
    year = "2023",
    address = "Singapore",
    publisher = "Association for Computational Linguistics",
    url = "https://aclanthology.org/2023.findings-emnlp.966/",
    doi = "10.18653/v1/2023.findings-emnlp.966",
}

@inproceedings{qwenmathprm,
    title = "The Lessons of Developing Process Reward Models in Mathematical Reasoning",
    author = "Zhang, Zhenru  and
      Zheng, Chujie  and
      Wu, Yangzhen  and
      Zhang, Beichen  and
      Lin, Runji  and
      Yu, Bowen  and
      Liu, Dayiheng  and
      Zhou, Jingren  and
      Lin, Junyang",
    editor = "Che, Wanxiang  and
      Nabende, Joyce  and
      Shutova, Ekaterina  and
      Pilehvar, Mohammad Taher",
    booktitle = "Findings of the Association for Computational Linguistics: ACL 2025",
    month = jul,
    year = "2025",
    address = "Vienna, Austria",
    publisher = "Association for Computational Linguistics",
    url = "https://aclanthology.org/2025.findings-acl.547/",
    doi = "10.18653/v1/2025.findings-acl.547",
    pages = "10495--10516",
    ISBN = "979-8-89176-256-5",
}

@article{sheng2024hybridflow,
  title   = {HybridFlow: A Flexible and Efficient RLHF Framework},
  author  = {Guangming Sheng and Chi Zhang and Zilingfeng Ye and Xibin Wu and Wang Zhang and Ru Zhang and Yanghua Peng and Haibin Lin and Chuan Wu},
  year    = {2024},
  journal = {arXiv preprint arXiv: 2409.19256}
}

@inproceedings{vllm,
author = {Kwon, Woosuk and Li, Zhuohan and Zhuang, Siyuan and Sheng, Ying and Zheng, Lianmin and Yu, Cody Hao and Gonzalez, Joseph and Zhang, Hao and Stoica, Ion},
title = {Efficient Memory Management for Large Language Model Serving with PagedAttention},
year = {2023},
isbn = {9798400702297},
publisher = {Association for Computing Machinery},
address = {New York, NY, USA},
url = {https://doi.org/10.1145/3600006.3613165},
doi = {10.1145/3600006.3613165},
booktitle = {Proceedings of the 29th Symposium on Operating Systems Principles},
pages = {611–626},
numpages = {16},
location = {Koblenz, Germany},
series = {SOSP '23}
}

@misc{gpt-4o,
  title = {Hello GPT-4o},
  author = {OpenAI},
  year = {2024},
  url = {https://openai.com/index/hello-gpt-4o/}
}

@article{llama3modelcard,
    title={Llama 3 Model Card},
    author={AI@Meta},
    year={2024},
    url = {https://github.com/meta-llama/llama3/blob/main/MODEL_CARD.md}
}

@misc{openai2026gpt54mini,
  title={Introducing {GPT-5.4} mini and nano},
  author={{OpenAI}},
  year={2026},
  month={March},
  howpublished={\url{https://openai.com/index/introducing-gpt-5-4-mini-and-nano/}},
  note={Accessed: 2026-05-26}
}

@misc{bge_m3,
      title={BGE M3-Embedding: Multi-Lingual, Multi-Functionality, Multi-Granularity Text Embeddings Through Self-Knowledge Distillation}, 
      author={Jianlv Chen and Shitao Xiao and Peitian Zhang and Kun Luo and Defu Lian and Zheng Liu},
      year={2024},
      eprint={2402.03216},
      archivePrefix={arXiv},
      primaryClass={cs.CL}
}

@misc{nomic_code,
      title={CoRNStack: High-Quality Contrastive Data for Better Code Retrieval and Reranking}, 
      author={Tarun Suresh and Revanth Gangi Reddy and Yifei Xu and Zach Nussbaum and Andriy Mulyar and Brandon Duderstadt and Heng Ji},
      year={2025},
      eprint={2412.01007},
      archivePrefix={arXiv},
      primaryClass={cs.CL},
      url={https://arxiv.org/abs/2412.01007}, 
}

@misc{jinacode,
      title={Jina Embeddings 2: 8192-Token General-Purpose Text Embeddings for Long Documents}, 
      author={Michael G{\"u}nther and Jackmin Ong and Isabelle Mohr and Alaeddine Abdessalem and Tanguy Abel and Mohammad Kalim Akram and Susana Guzman and Georgios Mastrapas and Saba Sturua and Bo Wang and Maximilian Werk and Nan Wang and Han Xiao},
      year={2024},
      howpublished={arXiv preprint arXiv:2310.19923},
      url={https://arxiv.org/abs/2310.19923},
      primaryClass={cs.CL}, 
}

@misc{openai2024embeddings,
  title        = {New embedding models and {API} updates},
  author       = {{OpenAI}},
  year         = {2024},
  howpublished = {\url{https://openai.com/index/new-embedding-models-and-api-updates/}},
  note         = {Accessed: 2026-05-26}
}

@misc{gsm8k,
      title={Training Verifiers to Solve Math Word Problems}, 
      author={Karl Cobbe and Vineet Kosaraju and Mohammad Bavarian and Mark Chen and Heewoo Jun and Lukasz Kaiser and Matthias Plappert and Jerry Tworek and Jacob Hilton and Reiichiro Nakano and Christopher Hesse and John Schulman},
      year={2021},
      eprint={2110.14168},
      archivePrefix={arXiv},
      primaryClass={cs.LG},
      url={https://arxiv.org/abs/2110.14168}, 
}

@misc{claude_sonnet_5,
  author       = {{Anthropic}},
  title        = {Claude Sonnet 5},
  year         = {2026},
  howpublished = {\url{https://www.anthropic.com/news/claude-sonnet-5}},
}

@misc{gpt_52,
  author       = {{OpenAI}},
  title        = {Introducing GPT-5.2},
  year         = {2025},
  howpublished = {\url{https://openai.com/ko-KR/index/introducing-gpt-5-2/}},
}

\nocite{vllm}

\appendix

\newpage
\section{Experimental Setup}

\subsection{Sampling and Verification}

\paragraph{Sampling prompt}
Unless otherwise specified, we use the sampling prompt in Figure~\ref{fig:solution_generation_prompt} for all solution generation. 
This includes both offline solution generation for analysis and rollout sampling during RLVR training and evaluation. 

\paragraph{Answer verification}
We verify final-answer correctness using the Qwen3-4B model with the prompt in Figure~\ref{fig:answer_verification_prompt}. 
For each generated solution, we extract the predicted answer from the \verb|\boxed{}| expression; if no boxed answer is found, we use the last 300 characters of the solution instead. The verifier judge is then given the predicted answer and the ground-truth answer, and is instructed to output either \texttt{correct} or \texttt{incorrect}.

We use a judge-based verifier following its robustness compared to rule-based matching for general mathematical answers, where equivalent answers may be written in different forms. 

{%
\begin{tcolorbox}[
  width=\linewidth,
  breakable,
  title=\textbf{Base Sampling Prompt},
  colback=gray!5,
  colframe=gray!75,
  boxrule=0.5pt,
  arc=3pt,
  left=6pt,
  right=6pt,
  top=6pt,
  bottom=6pt
]

\begin{Verbatim}[
  fontsize=\scriptsize,
  breaklines=true,
  breakanywhere=true,
  breaksymbolleft={},
  breaksymbolright={},
]
### Problem: {problem}
Please reason step by step, and put your answer within \boxed{}
### Solution:
\end{Verbatim}

\end{tcolorbox}
\captionof{figure}{Solution generation prompt.}
\label{fig:solution_generation_prompt}
}%

\vspace{1em}

{%
\begin{tcolorbox}[
  width=\linewidth,
  breakable,
  title=\textbf{Answer Verification Prompt},
  colback=gray!5,
  colframe=gray!75,
  boxrule=0.5pt,
  arc=3pt,
  left=6pt,
  right=6pt,
  top=6pt,
  bottom=6pt
]

\begin{Verbatim}[
  fontsize=\scriptsize,
  breaklines=true,
  breakanywhere=true,
  breaksymbolleft={},
  breaksymbolright={},
]
System message:
You are a math expert.
You are given a golden answer and a predicted answer from a solver.
You need to verify if the predicted answer is correct.
Only output "correct" or "incorrect".

Few-shot messages:
User:
Golden answer: 540, Predicted answer: The total number of ways the cars
can stack up so that all three lanes are occupied is calculated to be 750.

Assistant:
incorrect

User:
Golden answer: 3, Predicted answer: The ratio \frac{AC}{AE} = 3.

Assistant:
correct

Query message:
User:
Golden answer: {golden_answer}, Predicted answer: {predicted_answer}

Output constraint:
Assistant:
correct
or
incorrect
\end{Verbatim}
\end{tcolorbox}
\captionof{figure}{Answer verification prompt. The model is instructed to determine whether a predicted answer matches the golden answer, using few-shot examples and constrained to output either ``correct'' or ``incorrect''.}
\label{fig:answer_verification_prompt}
\leavevmode\par
}%


\subsection{Diversity Metrics under Analysis}
\label{sec:div_met_def}

Full list of analyzed diversity metrics and their detailed definitions are listed below:

\begin{itemize}
  \item \textbf{Cosine Embedding} given two solutions $s_A$, $s_B$, and an embedding model $E$, the distance between the two solutions is calculated as $d(s_A, s_B) = 1 - \frac{E(s_A) \cdot E(s_B)}{
  ||E(s_A)|| \cdot ||E(s_B)||}$
  \item \textbf{Self-BLEU} Proposed by \citet{zhu2018texygen}, Self-BLEU measures lexical homogeneity by treating each solution as a hypothesis against the rest as references:
    \[
    \mathrm{Self\text{-}BLEU}(S) = \frac{1}{N}\sum_{i=1}^{N} 
    \mathrm{BLEU}\!\left(s_i,\; S \setminus \{s_i\}\right).
    \]
    We report $1 - \mathrm{Self\text{-}BLEU}(S)$.
  \item \textbf{N-gram Distance} Given the $n$-gram multisets $G_n(s_A)$ and $G_n(s_B)$, the pairwise distance is defined as the Jaccard distance:
    \[
    d_n(s_A, s_B) = 1 - \frac{\left|G_n(s_A) \cap G_n(s_B)\right|}
    {\left|G_n(s_A) \cup G_n(s_B)\right|}.
    \]
  Our main analysis uses $n=2$. Ablations for different values of $n$ can be found in Appendix~\ref{app:detail_align_eval}
  \item \textbf{Distinct-Equations} (Distinct-EQ) \citet{wuprogress} The calculation of this metric is done in two steps: extracting equations from the input solution set and computing the ratio of unique equations. Let $\mathrm{Eq}(S)$ denote the multiset of all extracted equations from $S$. Then:
  \[
  D_{\mathrm{eq}}(S)=\frac{\left|\mathrm{Unique}\!\left(\mathrm{Eq}(S)\right)\right|}{\left|\mathrm{Eq}(S)\right|}
  \]
  \item \textbf{RPD} Suggested by \citet{ju2026reasoningpathdivergencenew},
  this metric aims to discriminate between two solutions for a math problem
  by their reasoning trajectory. Given two solutions $s_A$ and $s_B$, it first
  extracts step-level representations using an LLM:
  $L_A = a_1, a_2, \ldots, a_m$, $L_B = b_1, b_2, \ldots, b_n$.
  Each representation is then encoded into an embedding vector, yielding a
  step-level distance:
  \[
  d_i = \min_{j=1, \ldots, n} \left( 1 - \frac{e_{a_i}\cdot e_{b_j}}{\|e_{a_i}\| \cdot \|e_{b_j}\|} \right)
  \]
  The overall distance is calculated as the average of step-level distances
  \[
  d(s_A, s_B) = \frac{1}{m} \sum_{i=1}^m d_i
  \]
\end{itemize}

\subsection{Diversity-Aware RLVR Baselines}
\label{app:diversity_rlvr_baselines}

\paragraph{DQO.}
DQO~\citep{chen2026dqo} adds a group-level semantic diversity objective to RL post-training. For a prompt $x$, let $Y=\{y_1,\ldots,y_G\}$ be a group of responses and let $z_i=E(y_i)$ be the embedding of response $y_i$. DQO computes an embedding-similarity matrix $M(Y)$ over the group and uses its determinant as the diversity signal:
\[
D_{\mathrm{DQO}}(Y) = \log \det(M(Y) + \epsilon I).
\]
The objective can be summarized as
\[
\mathcal{J}_{\mathrm{DQO}}(\theta) = \mathbb{E}_{Y \sim \pi_\theta(\cdot \mid x)}
\left[
R(Y) + \alpha D_{\mathrm{DQO}}(Y)
\right],
\]
where $R(Y)$ denotes the quality reward and $\alpha$ controls the strength of the diversity objective. 

\paragraph{DIVER.}
DIVER~\citep{hu2026diver} adds an intrinsic diversity reward to the verifiable reward in RLVR. Given a group of rollouts $Y=\{y_1,\ldots,y_G\}$, each correct rollout receives a diversity bonus based on its dissimilarity to the other rollouts:
\[
\tilde r(x,y_i) = r_{\mathrm{ver}}(x,y_i) +
\beta \, \mathbf{1}\{r_{\mathrm{ver}}(x,y_i)=1\} D_i(Y),
\]
where $r_{\mathrm{ver}}$ is the correctness reward and $\beta$ controls the diversity strength. 

We experiment with two conventional diversity measures introduced in DIVER. Textual diversity (TD) computes $D_i(Y)$ from pairwise BLEU:
\[
D_i^{\mathrm{TD}}(Y) = \frac{1}{G-1}\sum_{j\neq i}\left(1-\mathrm{BLEU}(y_i,y_j)\right).
\]
Equation Diversity (ED) computes $D_i(Y)$ with the ratio of distinct equations:
\[
D_i^{\mathrm{ED}}(Y) = \frac{ \left|\mathrm{Eq}(y_i)\setminus \bigcup_{j\neq i}\mathrm{Eq}(y_j)\right|
}{ \max\{|\mathrm{Eq}(y_i)|,1\} }.
\]

\subsection{RLVR Training Details}

Unless otherwise stated, all diversity-aware RLVR experiments use the training hyperparameters in Table~\ref{tab:diversity_rlvr_hyperparams}. 
Method-specific hyperparameters not listed in the table are inherited from the official implementations.
For 3B models, we run all training on $2\times$ A100 GPUs. 
For 7B models, we use $4\times$ A100 GPUs for GRPO/DIVER-TD, and $2\times$ H200 GPUs for DQO/DIVER-ED.
We used the verl framework \citep{sheng2024hybridflow} for experiments.

For quality evaluation on OlympiadBench, we sample 32 solutions per problem with temperature 0.7, following common practice. For approach coverage analysis, we sample 64 solutions per problem with temperature 1.0 to better capture the range of approaches each policy can generate.

\begin{table}[t]
\centering
\small
\setlength{\tabcolsep}{8pt}
\renewcommand{\arraystretch}{1.05}
\begin{tabular}{lc}
\toprule
\textbf{Hyperparameter} & \textbf{Value} \\
\midrule
Learning Rate (LR) & $1 \times 10^{-6}$ \\
Optimizer & AdamW \\
Training Batch Size & 128 \\
PPO Mini-batch Size & 32 \\
Rollouts per Question ($N$) & 8 \\
KL Coefficient & 0.001 \\
Rollout Temperature & 1.0 \\
Max Input Tokens & 1024 \\
Max Response Tokens & 4096 \\
\bottomrule
\end{tabular}
\caption{Common hyperparameters used for diversity-aware RLVR training.}
\label{tab:diversity_rlvr_hyperparams}
\end{table}

\section{Human Annotation and LLM Judge}
\label{app:details_eval_framework}

\subsection{Details in Dataset Construction}
\label{app:dataset_construction}
For the main analysis in the paper, we selected problems guaranteed to have multiple solution approaches, as we observed that the number of unique and valid approaches to solving the problem was often constrained by the problem itself, e.g., when it is too simple or, conversely, very difficult. To this end, we introduce a four-stage filtering pipeline.

\begin{enumerate}
  \item Difficulty Filtering: Given a problem $x$, we generate 16 solutions using the Qwen3-4B model. In this step, we filter out problems with a pass rate below 0.2 and problems with an average solution length lower than 300 tokens. By this stage, we aim to filter out problems that are either too simple or, conversely, too difficult to produce reachable, valid approaches. 
  \item Plan Generation: Given a problem $x$, we prompt the GPT-5.2 model to generate four distinct candidate approaches, $C = \{c_1, c_2, c_3, c_4\}$. For this, we used the prompt in Figure~\ref{fig:strategy_enumeration_prompt}.
  \item Correctness Validation: For each $c \in C$, we prompt a smaller language model $\theta_{solve}$ to generate $N=8$ solutions from the conditioned distribution $p_{\theta_{solve}}(\cdot | x, c)$. We retain only those plans that yield at least one correct solution. By this, we aim to ensure both the correctness and 'reachability' of a specific approach - not only validating that this approach can lead to a correct answer, but also whether medium-sized language models, our primary generators in this paper, will be able to exploit such an approach effectively.
  \item Distinctness Judge: Among the filtered plans, the GPT-5.2 judge assesses whether the surviving plans constitute meaningfully distinct approaches, and problems with fewer than three such approaches are excluded. When tested on 50 validation problems with human labels, the LLM judge reported a precision of 0.9644, a recall of 0.8714, and an F1 score of 0.9156 across 5 independent runs.
\end{enumerate}

\noindent
We used the GPT-5.2 model for the concept generator $\theta_\mathrm{gen}$ and uniqueness judge $\mathcal{J}$, and the Qwen3-4B model with thinking mode as $\theta_\mathrm{solve}$. 
When applied to the MATH training dataset, this yields a training set of 2000 problems and an evaluation set of 469 problems; all analyses are conducted on the evaluation set unless otherwise specified. 

\paragraph{Cost analysis} The dataset construction stage (plan generation, correctness validation, and distinctness filtering over all 7,500 MATH training problems) cost approximately \$110 in GPT-5.2 Batch API calls. Coverage analysis (clustering up to 64 solutions for each of 150 problems) costs approximately \$7 per policy evaluated.

\subsection{Constructing Human Annotation Set}
\label{app:human_annotation}

Candidate problems were drawn from the multi-approach problem set 
of Section~\ref{sec:multi_approach_problems}. For each candidate, we 
generated 32 solutions per problem from each of three base models: 
Qwen3-8B-Base, Qwen2.5-32B-Base, and OLMo3-32B-Base.

We then constructed annotation items in two steps. First, we manually filtered the generated solutions, keeping only those correct in both the final answer and the intermediate reasoning. 
Second, for each problem, the authors inspected the remaining solutions and assigned approach labels, retaining the problem only if two same-approach pairs and two different-approach pairs could be formed from a single model's solutions. 
These author-provided labels were used solely to balance the annotation set and were not shown to annotators.

Problems were drawn from three problem types in the MATH dataset—Intermediate Algebra, Geometry, and Counting \& Probability—using the type labels provided by the dataset. Inter-annotator agreement varied notably across types: 91\% for Intermediate Algebra versus 67\% for both Geometry and Counting \& Probability. Further investigating how agreement varies across problem types, as well as across annotators with different mathematical backgrounds, would be an interesting direction for future work.

Annotators were recruited from graduate students and participated voluntarily without monetary compensation. 
Before annotation, they 
were informed that their labels would be used for ML/NLP research on evaluating diversity in LLM-generated mathematical reasoning. The exact instruction is shown in Figure~\ref{fig:human_annotation_prompt}.

\begin{figure*}[t]
\centering

{%
\begin{tcolorbox}[
  width=\textwidth,
  title=\textbf{Human Annotation Instructions},
  colback=gray!5,
  colframe=gray!75,
  boxrule=0.5pt,
  arc=3pt,
  left=6pt,
  right=6pt,
  top=6pt,
  bottom=6pt
]

\begin{Verbatim}[
  fontsize=\scriptsize,
  breaklines=true,
  breakanywhere=true,
  breaksymbolleft={},
  breaksymbolright={},
]
Instructions

For each item, you will see one problem and two correct solutions, Solution A and Solution B.

Your task. Decide whether the two solutions use the same high-level mathematical approach.

When judging the approach, consider:

- Mathematical tools: the core techniques used to solve the problem,
  such as algebraic vs. geometric methods.

- Structural definitions: the variables, objects, or constructions introduced
  to frame the solution, such as defining an auxiliary function vs. direct substitution.

- Representational viewpoint: the conceptual frame used to interpret the problem,
  such as coordinate geometry vs. synthetic geometry.

Choose one of:
- Same approach
- Different approach


Problem:
[Problem statement here]

Solution A:
[First correct solution here]

Solution B:
[Second correct solution here]
\end{Verbatim}

\end{tcolorbox}
}%

\caption{\textbf{Human annotation prompt.} Instructions and item format shown to annotators for pairwise approach judgments. Each item contains one problem and two correct model-generated solutions, and annotators decide whether the two solutions use the same high-level mathematical approach.}
\label{fig:human_annotation_prompt}

\end{figure*}

\subsection{Details of the LLM Judge Prompt}
\label{app:llm_judge_prompt}

This section provides the full prompt template used for the LLM judge in Section~\ref{sec:approach_judge_design}, and explains the design principles behind it.

\paragraph{Prompt design principles.}
The prompt was designed to enforce a conservative decision boundary.
First, it explicitly defines different approaches in terms of their underlying mathematical mechanisms and conceptual interpretations.
Second, it instructs the judge to merge solutions when a distinction is ambiguous, weakly supported, or mainly about presentation.
Third, it requires reusable outputs for downstream analysis.

In preliminary trials, we found that naively prompting the judge to cluster different approaches often led to over-splitting: solutions using the same mathematical idea were sometimes separated due to notation, formatting, verification steps, or equivalent reformulations.
The final prompt, therefore, includes an explicit negative list of surface-level differences that should not define a new approach, as well as a final verification step that asks the judge to merge groups that differ only in execution details.
Overall, the prompt separates solutions only when their main mathematical mechanism, structural definitions, or representational viewpoint clearly differs; otherwise, it merges solutions that differ only in presentation. All analyses were performed using the OpenAI Batch API with default sampling parameters. We set the reasoning effort to \texttt{None}, as we observed minimal performance gain with more reasoning budget. 

\paragraph{Prompt development}
We developed the judge prompt using a separate validation set with only author-provided labels, constructed independently of the human annotation set in Section~\ref{sec:human_anno}.
The human annotation set was not used during prompt development and was reserved only for the final alignment evaluation.

\subsection{Clustering procedure}
Given a list of correct solutions, we perform clustering in two stages using the same clustering prompt throughout.
We first split the solutions into chunks of at most eight and apply the LLM judge to each chunk independently.
We then select one representative from each intermediate cluster and run a final clustering pass over the representatives, without further chunking.
Finally, we assign each original solution to the final cluster of its representative.

Figure~\ref{fig:judge_real_world_case} presents an unedited real-world clustering example from our evaluation pipeline.

\begin{figure*}[p]
\centering

{%
\begin{tcolorbox}[
  width=\textwidth,
  title=\textbf{LLM Judge Prompt for Approach Clustering},
  colback=gray!5,
  colframe=gray!75,
  boxrule=0.5pt,
  arc=3pt,
  left=6pt,
  right=6pt,
  top=6pt,
  bottom=6pt
]

\begin{Verbatim}[
  fontsize=\scriptsize,
  breaklines=true,
  breakanywhere=true,
  breaksymbolleft={},
  breaksymbolright={},
]
You are an expert Mathematician specializing in the comparative analysis of
problem-solving strategies.
Your task is to evaluate a set of solutions, cluster them based on their
conceptual and mathematical distinctions, and provide a descriptive summary
for each cluster.

### DEFINITION OF DIFFERENT APPROACHES (STRICT)
You must apply the following definition strictly. Focus on the mechanism,
not surface features.

"When determining whether two solutions represent the same or different
approaches, focus on the underlying mathematical mechanism AND the conceptual
interpretation used in the reasoning."

Two solutions must be classified as DIFFERENT approaches if they rely on:
1. Different Mathematical Tools:
   (e.g., Calculus vs. Geometry vs. Number Theory).
2. Different Definitions/Structures:
   (e.g., Explicit formula vs. Recurrence relation).
3. Different Representational Viewpoints:
   (e.g., Geometric locus vs. Vector algebra; Slope as ratio vs.
   Trigonometric angle).

### CONSERVATIVE DECISION POLICY
The three criteria above define valid reasons for distinguishing approaches,
but you should create a separate group only when the difference is substantial,
central, and clearly changes the main proof route.

When the distinction is ambiguous, weakly supported, or mostly about
presentation, prefer merging rather than splitting.

If two solutions can be summarized by the same one-sentence explanation of why
the method works, they should usually be placed in the same group.

Do not create a separate group for differences that are only about notation,
variable names, order of steps, level of detail, algebraic cleanup,
verification steps, or equivalent reformulations of the same core idea.

### OUTPUT INSTRUCTIONS
For each identified group, you must provide:
1. Group Name:
   A concise technical label for the approach.
2. Core Idea:
   A 1-2 sentence plain-text explanation of the underlying mechanism.
   Explain what mathematical concept is the driver and how it frames the
   problem.
3. Solution IDs:
   The list of solution numbers belonging to this group.

Place each solution in only one approach group.

### FORMAT REQUIREMENTS
Output exactly one JSON object and nothing else.
Do not use Markdown code fences.
Do not use LaTeX, backslashes, or escaped math notation in any string field.
Use short plain-text strings only.
Use an ASCII snake_case style label for each group_name.
reasoning_trace must be 1-2 short plain-text sentences.

### FINAL VERIFICATION
Before finalizing, review every pair of groups and ask:
"Is the difference here about the core mathematical mechanism, or just about
execution details?"

If two groups use the same mathematical tool, structure, and viewpoint, merge
them, even if their step-by-step procedures look different.

### OUTPUT FORMAT
{
  "reasoning_trace": "(Brief overall analysis of how the solutions differ conceptually...)",
  "groups": [
    {
      "group_name": "...",
      "core_idea": "...",
      "solution_ids": [1, 3]
    },
    {
      "group_name": "...",
      "core_idea": "...",
      "solution_ids": [2]
    }
  ]
}
\end{Verbatim}

\end{tcolorbox}
}%

\caption{LLM judge prompt for approach clustering.}
\label{fig:prompt_llm_judge}

\end{figure*}

\subsection{Cross-model Validation}
Our suggested framework depends on the use of LLM-based system, especially on problem filtering and approach clustering. In our work, we adopt GPT-5.2 as the main backbone model due to its high reasoning and instruction-following ability. 
However, relying on a single judge model can produce judge-biased results. Following this concern, we reran the judge-dependent components using an independent Claude Sonnet 5 \citep{claude_sonnet_5} and re-evaluated the downstream results.

\paragraph{Problem Filtering}
First, we cross-validate GPT-5.2 and Claude Sonnet~5 as interchangeable plan generators and distinctness judges against human labels (Table~\ref{tab:cross_judge_filtering}). Precision---the critical quantity for dataset soundness---remains high (0.87--0.98) across all configurations, and swapping the judge on a fixed plan set changes the final decision only marginally (0.914 vs.\ 0.865). This indicates that the judgement of approach distinctness is not an artifacts of a particular model, but is applicable across the two widely-used proprietary models.

\begin{table}[h]
\centering
\begin{tabular}{lccc}
\toprule
Plans $\times$ Judge & P & R & F1 \\
\midrule
GPT $\times$ GPT       & 0.980 & 0.857 & 0.914 \\
GPT $\times$ Claude    & 0.938 & 0.804 & 0.865 \\
Claude $\times$ Claude & 0.865 & 0.804 & 0.833 \\
Claude $\times$ GPT    & 0.930 & 0.714 & 0.808 \\
\bottomrule
\end{tabular}
\caption{Problem filtering results across different plan generator and distinctness judge combinations.}
\label{tab:cross_judge_filtering}
\end{table}

\section{Analyses of Conventional Metrics}

\subsection{Scaffolding Ratio}
\label{app:scaffolding_ratio}

We measure shared scaffolding with a soft unigram-overlap ratio. 
For each problem $x$, let $S_x=\{s_1,\ldots,s_n\}$ be all correct solutions.
We lowercase each solution and tokenize it by whitespace. 
For each token $w$, we first compute its prevalence across correct solutions:
\[
p_x(w)=\frac{1}{n}\sum_{i=1}^{n}\mathbf{1}[w\in U_i],
\]
where $U_i$ is the set of unique tokens in solution $s_i$.
The scaffolding ratio of a solution is then the average prevalence of its
tokens:
\[
r(s_i)=\frac{1}{|T_i|}\sum_{w\in T_i} p_x(w),
\]
where $T_i$ is the token sequence of $s_i$.
Finally, the problem-level scaffolding ratio is
\[
\mathrm{Scaffold}(x)=\frac{1}{n}\sum_{i=1}^{n} r(s_i).
\]
\noindent
This score is high when a large fraction of solution tokens also appear in many other correct solutions to the same problem, capturing shared setup, repeated notation, restated conditions, and answer-formatting patterns.

\subsection{Detailed Concordance Implementation}
\label{app:detail_align_eval}

For each problem and model, we sample $N=32$ solutions at temperature 
$T=1.0$, keep only correct ones, and cluster them into approach groups 
using the LLM judge. For each tier $t$ and subset size $k$, we 
construct comparison pairs $(S_{\mathrm{high}}, S_{\mathrm{low}})$ of 
size $k$, where $S_{\mathrm{low}}$ covers $t$ distinct approaches and 
$S_{\mathrm{high}}$ covers $t+1$, with solutions allocated as evenly 
as possible across the selected approaches within each set. We retain 
a problem for a given $(t,k)$ setting only if at least $B=500$ valid 
pairs can be constructed; Table~\ref{tab:ccc_feasible_counts} reports 
the resulting counts. Higher-tier settings yield fewer feasible 
problems, but each retained problem still contributes $B=500$ pair 
comparisons.

\begin{table}[t]
\centering
\small
\begin{tabular}{llccc}
\toprule
Tier & $k$ & Qwen2.5-32B & Qwen3-8B & OLMo3-32B \\
\midrule
Tier 1 & 4 & 102 & 178 & 130 \\
Tier 1 & 6 & 59  & 131 & 93  \\
\arrayrulecolor{gray!65}\midrule[0.4pt]\arrayrulecolor{black}
Tier 2 & 4 & 92  & 112 & 96  \\
Tier 2 & 6 & 35  & 64  & 47  \\
\arrayrulecolor{gray!65}\midrule[0.4pt]\arrayrulecolor{black}
Tier 3 & 4 & 57  & 58  & 60  \\
Tier 3 & 6 & 30  & 43  & 35  \\
\bottomrule
\end{tabular}
\caption{Number of feasible problems used in the concordance evaluation for each tier, subset size $k$, and generator model.}
\label{tab:ccc_feasible_counts}
\end{table}

\begin{figure*}[t]
  \centering
  \includegraphics[width=\linewidth]{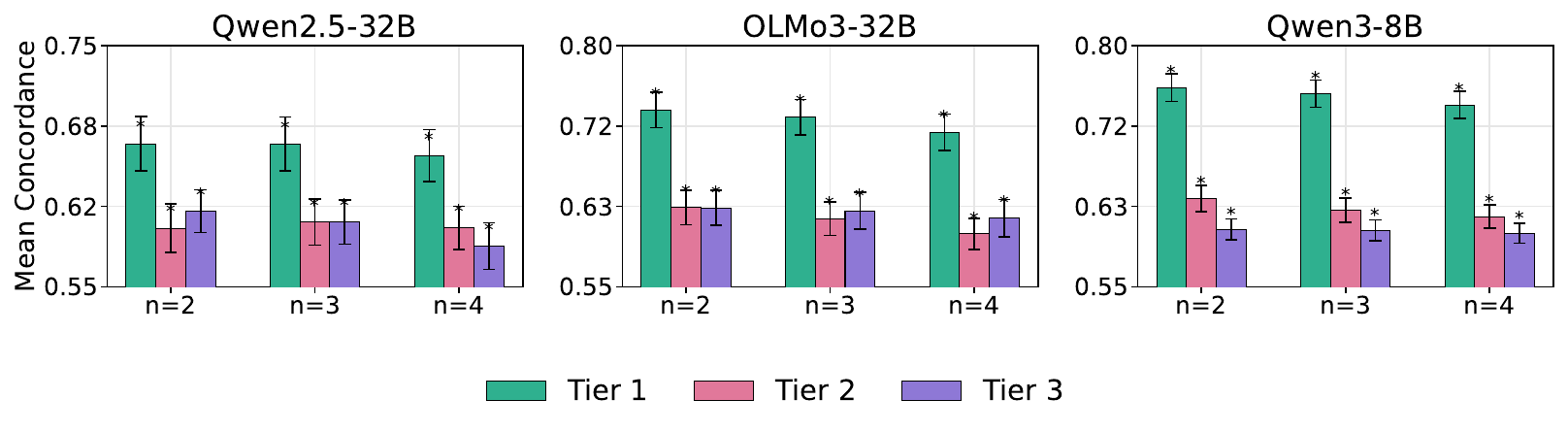}
  \caption{Effect of n-gram size on alignment with approach-level distance. Smaller n yields stronger alignment, with n=2 performing best among n $\in$ {2, 3, 4}.}
  \label{fig:ngram_ablations}
\end{figure*}
\begin{figure*}[t]
  \centering
  \includegraphics[width=\linewidth]{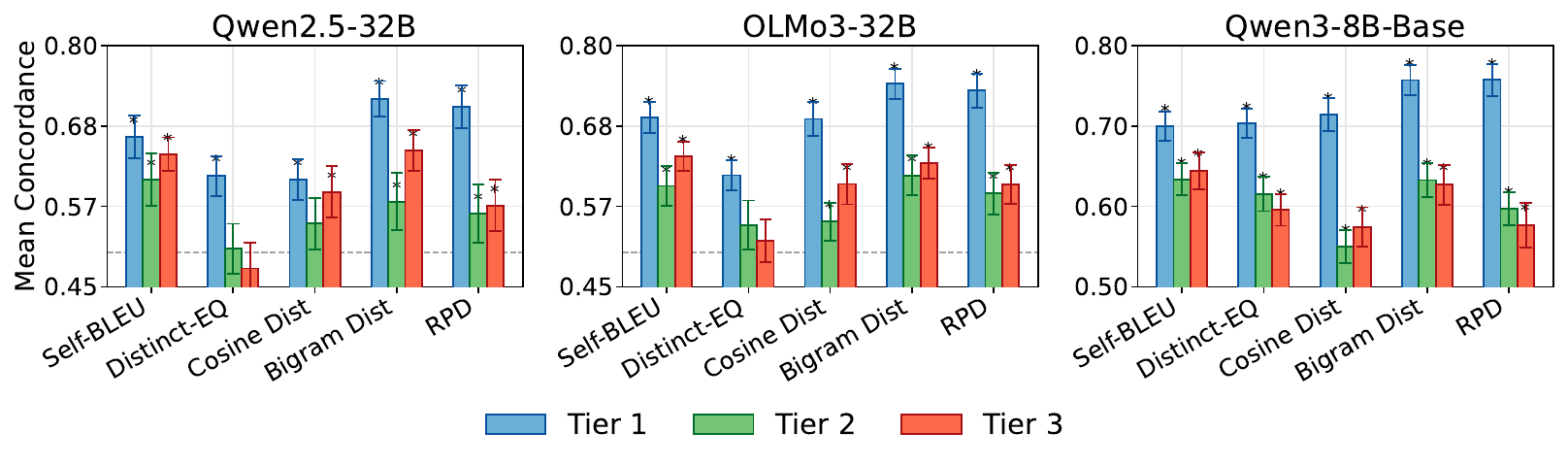}
  \caption{Set-level concordance ($C_{\mathrm{set}}$) at subset size $k=6$ across solver models. Metrics are compared on their ability to rank higher-approach-diversity subsets above matched lower-diversity subsets.}
  \label{fig:ccc_k6_result}
\end{figure*}
\begin{figure*}[t]
  \centering
  \includegraphics[width=\linewidth]{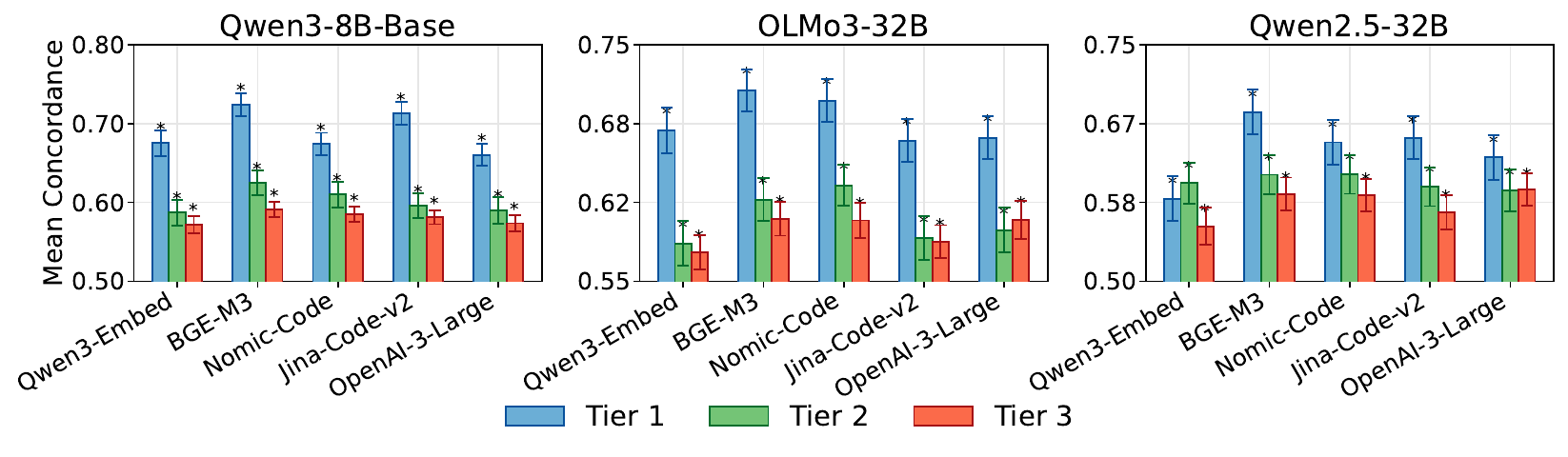}
  \caption{Set-level concordance ($C_{\mathrm{set}}$) at subset size $k=4$ across different embedding models.}
  \label{fig:embedding_ablations}
\end{figure*}
\begin{figure*}[t]
  \centering
  \includegraphics[width=\linewidth]{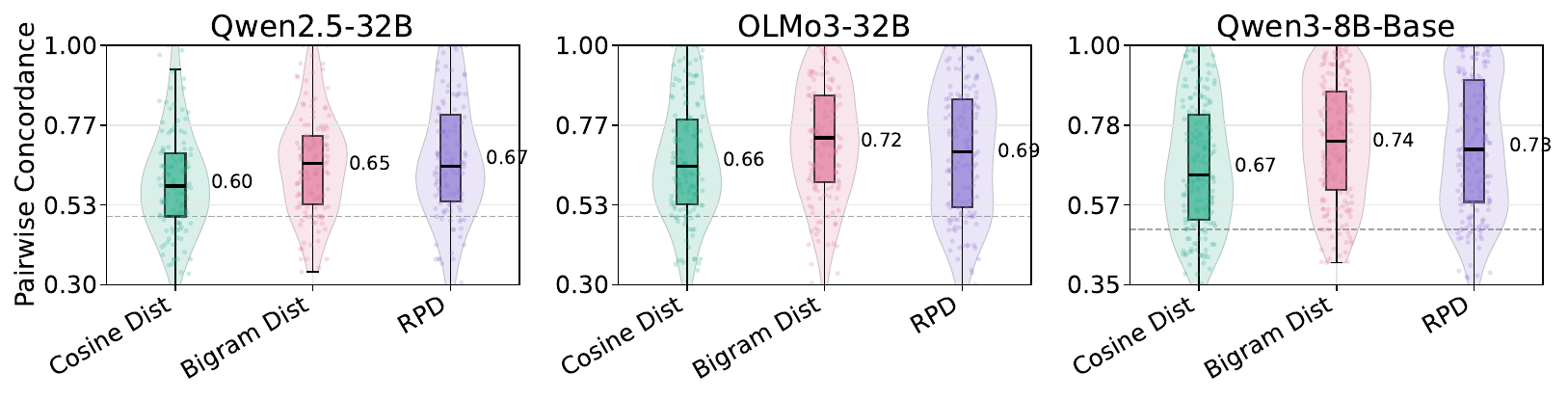}
  \caption{Pairwise concordance ($C_{\mathrm{pair}}$) of N-gram distance, cosine distance, and RPD across solver models. Higher is better, indicating stronger alignment with approach-level differences.}
  \label{fig:pairwise-multimodel}
\end{figure*}

\subsection{Ablations on Conventional Measures}
We report ablations on conventional measures by varying the N-gram order $n$ (Figure~\ref{fig:ngram_ablations}), the embedding backbone for cosine distance (Figure~\ref{fig:embedding_ablations}), and the subset size $k$ in set-level concordance evaluation (Figure~\ref{fig:ccc_k6_result}). Results are consistent with the main paper: across tiers 2 and 3, all metrics fail to reliably capture approach-level diversity beyond coarse surface differences.

\paragraph{Embedding model ablation}
We test whether the weakness of cosine distance depends on the embedding backbone.
We compare two general-purpose models, BAAI/bge-m3~\citep{bge_m3} and the OpenAI embedding API~\citep{openai2024embeddings}, with two code-oriented models, nomic-ai/nomic-embed-code~\citep{nomic_code} and jina-embeddings-v2-base-code~\citep{jinacode}.
We include the latter as long-context, domain-specialized alternatives, since long-context math-specific embedding models are limited.
Figure~\ref{fig:embedding_ablations} shows that the conclusion is unchanged: cosine distance still degrades in Tiers 2 and 3.

\subsection{Pairwise Distance Evaluation}
The set-level failure observed in the previous section could, in principle, stem from aggregation rather than the metrics themselves. To isolate this, we evaluate pairwise distances directly. For each problem $x$, we form same-approach pairs $\mathcal{P}_{\mathrm{same}}(x)$ and different-approach pairs $\mathcal{P}_{\mathrm{diff}}(x)$, and measure the pairwise concordance from Equation~\ref{eq:pairwise_concordance}.

\begin{equation}
\label{eq:pairwise_concordance}
\begin{split}
C_{\mathrm{pair}}(d) = \frac{1}{B}\sum_{b=1}^{B} \mathds{1}\!\Bigl[\,
& d\!\left((s_i, s_j)_{\mathrm{diff}}^{(b)}\right) \\
& > d\!\left((s_i, s_j)_{\mathrm{same}}^{(b)}\right) \Bigr]
\end{split}
\end{equation}

Figure~\ref{fig:pairwise-multimodel} mirrors the set-level ranking: cosine distance is weakest across all three models, N-gram distance strongest in two of three, and RPD — despite its LLM-based summarization and step decomposition — does not surpass simple bigram overlap. 
We also note that pairwise concordance falls within a similar range to the Tier~1 set-level results, where surface and approach diversity are most easily separable, indicating the failure lies in the distance metrics themselves rather than in the set-level aggregation. 
Per-problem variance is also substantial, indicating that the pairwise metrics often struggle to discriminate between same- and different-approach solution pairs. 

\subsection{Analysis on Additional Math Reasoning Benchmarks}
\begin{figure}[t]
  \centering
  \includegraphics[width=\linewidth]{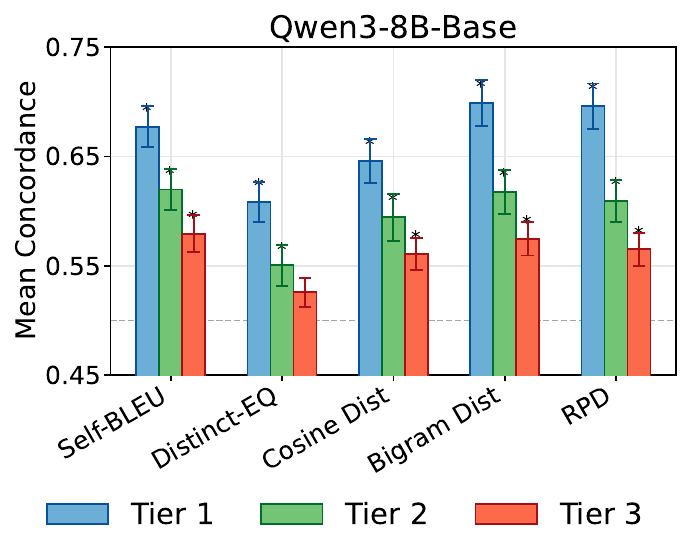}
  \caption{Set-level concordance ($C_{\mathrm{pair}}$) of N-gram distance, cosine distance, and RPD on OlympiadBench-sourced problems.}
  \label{fig:ccc_olympiadbench}
\end{figure}

To evaluate whether our findings hold beyond MATH-sourced problems, we apply our approach-feasibility filtering pipeline to two widely used benchmarks of differing difficulty: easy GSM8K \citep{gsm8k} and harder OlympiadBench. The proportion of multi-approach-feasible problems rises monotonically with difficulty: 0.7\% on GSM8K (9/1319), 32.9\% on the MATH training set (2469/7500), and 48.6\% on OlympiadBench (328/675), indicating that approach-level diversity becomes more prevalent on harder benchmarks.

As GSM8K yields only nine problems after filtering, we conduct the additional analysis on OlympiadBench only, following the procedure of the main paper ($N=32$ solutions from Qwen3-8B-Base, temperature 1.0, subset size $k=4$). The results in Figure~\ref{fig:ccc_olympiadbench} mirror our main results in Figure~\ref{fig:ccc_k4_result}: all metrics retain reasonable discriminative power in Tier~1, while concordance drops substantially in Tiers~2 and~3, with the same qualitative ordering of metrics. 

This support the generalizability of our findings in two respects: (i) the surface--approach gap persists on other mathematical benchmarks, and (ii) the fraction of multi-approach-feasible problems grows with difficulty, suggesting that our framework is applicable to more challenging settings.

\section{Diversity-aware RLVR}
\label{app:details_diversity_RLVR}

\subsection{cov@N Estimator}
\label{app:coverage}

In Section~\ref{sec:diversity_rl}, we use $\mathrm{cov}_x(N, \pi)$,
the expected number of distinct approaches observed when sampling $N$
correct responses from a policy $\pi$ for a problem $x$, as our primary
measure of approach-level diversity:
\begin{equation}
    \mathrm{cov}_x(N, \pi)
    \;\coloneqq\; \mathbb{E}_{S_x \sim \pi(\cdot \mid x)}\!
        \left[\,\bigl|\mathcal{J}(S_x)\bigr|\,\right],\
\end{equation}
where $S_x$ is a set of $N$ correct solutions and $\mathcal{J}(S_x)$
denotes the set of approach clusters from $S_x$.

\begin{table*}[t]
    \centering
    \setlength{\tabcolsep}{10pt}
    \renewcommand{\arraystretch}{1.15}

    \begin{tabular}{lccc}
        \toprule
        & \multicolumn{2}{c}{Quality}
        & \multicolumn{1}{c}{Diversity} \\
        \cmidrule(lr){2-3}
        \cmidrule(lr){4-4}

        & Pass@1 & Pass@32 & Cov@16 \\
        \midrule

        Base
        & 15.19
        & 60.30
        & 2.302 \\

        \midrule

        GRPO
        & $28.44 \pm 0.79$
        & $65.13 \pm 0.62$
        & $1.707 \pm 0.091$ \\

        DQO
        & $25.78 \pm 0.65$
        & $62.96 \pm 0.78$
        & $1.690 \pm 0.141$ \\

        DIVER-TD
        & $28.75 \pm 0.56$
        & $66.47 \pm 0.82$
        & $1.613 \pm 0.088$ \\

        DIVER-ED
        & $29.31 \pm 0.25$
        & $65.23 \pm 1.25$
        & $1.753 \pm 0.124$ \\

        \bottomrule
    \end{tabular}
    \caption{Quality and diversity comparison across different diversity-aware RL algorithms, trained on Qwen2.5-3B model. Training was done across three different random seeds, reported values are mean ± std over seeds.}
    \label{tab:div_rl_seed_var}
\end{table*}
\begin{table}[t]
    \centering
    \resizebox{\columnwidth}{!}{%
        \begin{tabular}{lccc}
            \toprule
            & \multicolumn{2}{c}{Quality}
            & Diversity \\
            \cmidrule(lr){2-3}
            \cmidrule(lr){4-4}

            Method
            & Pass@1
            & Pass@32
            & Cov@16 \\
            \midrule

            GRPO
            & 31.39
            & 63.70
            & 1.450 \\

            DQO
            & 25.87
            & 64.74
            & 1.587 \\

            DIVER-TD
            & 32.37
            & 66.22
            & 1.453 \\

            \bottomrule
        \end{tabular}%
    }
    \caption{Quality and diversity comparison across different diversity-aware RL algorithms, trained for 200 steps with Qwen2.5-3B model.}
    \label{tab:div_rl_long_train}
\end{table}

\paragraph{Estimation}
For each problem $x$ we draw rollouts from $\pi(\cdot \mid x)$, and collect only correct solutions $S^\star_x$ with $|S^\star_x| = M \ge N$. Clustering $S^\star_x$ at the approach level yields a partition
\begin{equation}
    \mathcal{C}(x) \;=\; \{C_1, C_2, \dots, C_k\},
    \quad n_i \coloneqq |C_i|
\end{equation}

We can then estimate $\mathrm{cov}_x(N, \pi)$ by following the procedure below. Let $\mathbb{I}_i$ denote the
indicator that cluster $C_i$ is represented in such a subsample; cluster
$C_i$ is absent iff all $N$ items are drawn from the $M - n_i$
solutions outside $C_i$, so
\begin{equation}
    \Pr[\mathbb{I}_i = 0]
    \;=\; \frac{\binom{M - n_i}{N}}{\binom{M}{N}}.
\end{equation}
By linearity of expectation, this yields the unbiased estimator:
\begin{equation}
    \widehat{\mathrm{cov}}_x(N, \pi)
    \;=\; \mathbb{E}\!\left[\sum_{i=1}^{k} \mathbb{I}_i\right]
    \;=\; \sum_{i=1}^{k}\!\left(
        1 - \frac{\binom{M - n_i}{N}}{\binom{M}{N}}
      \right).
\end{equation}

\noindent
We set $N = 16$ and $M = 64$, and restrict the analysis to problems for which every checkpoint of Qwen2.5-3B models---including the base model---yields at least $N$ correct solutions, ensuring that the estimate is comparable across all checkpoints.

\subsection{DIVER Gain Decomposition}
\label{app:diver_analysis_detail}

Section~\ref{sec:diversity_rl} shows that preserving a target diversity metric does not necessarily preserve approach diversity. 
For DIVER, the recovered textual and equation-level diversity mainly reflects greater variation within the same approach, not broader coverage of distinct approaches. We provide the detailed measurement and decomposition below.

\paragraph{Textual diversity}
To understand what the increase in TD actually represents, we decompose TD into three parts. Let $\alpha_t$ be the fraction of same-approach pairs, and let $\bar{\alpha}$ denote its average over the two checkpoints being compared.
Also, $\mu_t^{\mathrm{intra}}$ and $\mu_t^{\mathrm{inter}}$ are the average distances within and across approaches. Then
\[
TD_t = \alpha_t \mu_t^{\mathrm{intra}}
+ (1-\alpha_t)\mu_t^{\mathrm{inter}}.
\]
The change in TD between two checkpoints is
\[
\begin{aligned}
\Delta \mathrm{TD}
={}& \bar{\alpha}\Delta\mu^{\mathrm{intra}}
+ (1-\bar{\alpha})\Delta\mu^{\mathrm{inter}} \\
&+ \Delta\alpha
(\bar{\mu}^{\mathrm{intra}}-\bar{\mu}^{\mathrm{inter}}).
\end{aligned}
\]
Here, the first term captures changes in distances among same-approach pairs, the second term captures changes in distances among different-approach pairs, and the third term captures changes in the pair composition. 
This decomposition allows us to identify where the increase in TD stems.

\paragraph{Equation diversity}
We analyze the ED gain by comparing two types of four-solution subsets: same-approach subsets, where all solutions share the same approach, and mixed-approach subsets, where two solutions are drawn from each of two distinct approaches. 
ED increases mainly for same-approach subsets, while remaining relatively stable for mixed-approach subsets. This suggests that the ED gain mostly reflects greater equation-level variation within the same approach.

\subsection{Robustness of the RLVR Findings}

\paragraph{Training seeds}
To account for the variance in RLVR training, we retrain every RLVR method on Qwen2.5-3B with three random seeds and report the results in Table~\ref{tab:div_rl_seed_var}. The results are consistent with our main findings: every algorithm fails to preserve the approach coverage of the base model, with a gap of roughly 4--7 standard deviations of the seed distribution. Moreover, the mean $\pm$ std intervals of approach coverage for the diversity-aware methods overlap with that of vanilla GRPO, again showing that optimizing a surface-level diversity objective does not preserve approach-level diversity.

\paragraph{Training length}
We further examine how diversity-aware RLVR methods behave under longer training. We train Qwen2.5-3B with DQO and DIVER-TD for 200 training steps and report the results in Table~\ref{tab:div_rl_long_train}. Although DQO retains relatively higher approach coverage at step 200, it gains essentially no accuracy gains from the extra 100 training steps. Moreover, regardless of whether a method is diversity-aware, approach diversity continues to decline as training progresses.

\paragraph{Judge model}
Next, we evaluate the approach clustering results with the Claude judge. When evaluated against the human-annotations in Section~\ref{sec:human_anno}, the Claude judge matches the GPT judge's overall agreement with human annotations (85\%), and agrees on 89.1\% of the 64 unambiguous cases where all annotators concur (vs.\ 96.9\% for GPT). 

We then re-examined cov@16 for the Qwen2.5-3B checkpoints in Section~\ref{sec:diversity_rl}. The results in Table~\ref{tab:cross_judge_cov}, show that both central findings hold under either judge: (1)~every RLVR-trained model falls below the base model, and (2)~no diversity-aware method significantly preserves coverage relative to naive GRPO.

\begin{table}[h]
\centering
\begin{tabular}{lcc}
\toprule
Method & GPT judge & Claude judge \\
\midrule
Base       & 2.302 & 2.129 \\
Naive GRPO & 1.786 & 1.791 \\
DQO        & 1.853 & 1.766 \\
DIVER-TD   & 1.519 & 1.756 \\
DIVER-ED   & 1.884 & 1.712 \\
\bottomrule
\end{tabular}
\caption{cov@16 of Qwen2.5-3B checkpoints introduced in Section~\ref{sec:diversity_rl} under two judge models, GPT-5.2 and Claude-Sonnet-5}
\label{tab:cross_judge_cov}
\end{table}

\section{Test-Time Scaling and Judge-Reward RLVR}

\subsection{Approach-Controlled Test-Time Scaling}
\label{app:test_time_scaling_details}

\begin{table*}[t]
\centering
\setlength{\tabcolsep}{3pt}
\renewcommand{\arraystretch}{1.15}

\resizebox{\textwidth}{!}{%
\begin{tabular}{c c c|ccc|ccc|ccc}
\toprule
& & &
\multicolumn{3}{c|}{Qwen2.5-3B} &
\multicolumn{3}{c|}{Llama-3.2-3B} &
\multicolumn{3}{c}{GPT-4o-mini} \\
\cmidrule(lr){4-6}
\cmidrule(lr){7-9}
\cmidrule(lr){10-12}

Strategy & & $M$
& $u=1$ & $u=2$ & $u=4$
& $u=1$ & $u=2$ & $u=4$
& $u=1$ & $u=2$ & $u=4$ \\
\midrule


& & 4
& \makecell{51.0 \\ {\scriptsize [47.7, 54.3]}}
& \makecell{54.1 \\ {\scriptsize [50.7, 57.5]}}
& \makecell{56.6 \\ {\scriptsize [53.0, 60.2]}}
& \makecell{36.4 \\ {\scriptsize [33.6, 39.2]}}
& \makecell{40.6 \\ {\scriptsize [37.5, 43.7]}}
& \makecell{45.6 \\ {\scriptsize [42.1, 49.1]}}
& \makecell{60.7 \\ {\scriptsize [57.4, 64.0]}}
& \makecell{64.1 \\ {\scriptsize [60.7, 67.5]}}
& \makecell{66.0 \\ {\scriptsize [62.5, 69.5]}} \\

Majority voting & & 8
& \makecell{55.6 \\ {\scriptsize [52.2, 59.1]}}
& \makecell{59.9 \\ {\scriptsize [56.3, 63.5]}}
& \makecell{61.8 \\ {\scriptsize [58.0, 65.5]}}
& \makecell{41.3 \\ {\scriptsize [38.2, 44.4]}}
& \makecell{47.3 \\ {\scriptsize [43.9, 50.7]}}
& \makecell{52.4 \\ {\scriptsize [48.7, 56.1]}}
& \makecell{64.4 \\ {\scriptsize [60.9, 67.8]}}
& \makecell{67.6 \\ {\scriptsize [64.1, 71.1]}}
& \makecell{69.5 \\ {\scriptsize [65.9, 73.1]}} \\

& & 16
& \makecell{58.1 \\ {\scriptsize [54.6, 61.7]}}
& \makecell{62.9 \\ {\scriptsize [59.2, 66.6]}}
& \makecell{64.9 \\ {\scriptsize [61.1, 68.8]}}
& \makecell{45.0 \\ {\scriptsize [41.6, 48.4]}}
& \makecell{51.4 \\ {\scriptsize [47.8, 55.0]}}
& \makecell{57.4 \\ {\scriptsize [53.5, 61.3]}}
& \makecell{65.2 \\ {\scriptsize [61.6, 68.8]}}
& \makecell{69.6 \\ {\scriptsize [66.1, 73.2]}}
& \makecell{71.6 \\ {\scriptsize [67.9, 75.2]}} \\

\midrule


& & 4
& \makecell{58.9 \\ {\scriptsize [55.6, 62.1]}}
& \makecell{61.9 \\ {\scriptsize [58.6, 65.3]}}
& \makecell{63.7 \\ {\scriptsize [60.2, 67.1]}}
& \makecell{40.5 \\ {\scriptsize [37.4, 43.5]}}
& \makecell{45.2 \\ {\scriptsize [42.0, 48.5]}}
& \makecell{48.9 \\ {\scriptsize [45.3, 52.4]}}
& \makecell{66.3 \\ {\scriptsize [63.1, 69.5]}}
& \makecell{70.1 \\ {\scriptsize [66.9, 73.3]}}
& \makecell{71.8 \\ {\scriptsize [68.5, 75.0]}} \\

Best-of-$N$ (PRM) & & 8
& \makecell{63.7 \\ {\scriptsize [60.4, 67.0]}}
& \makecell{67.2 \\ {\scriptsize [63.9, 70.5]}}
& \makecell{68.6 \\ {\scriptsize [65.2, 72.0]}}
& \makecell{43.7 \\ {\scriptsize [40.5, 46.9]}}
& \makecell{48.8 \\ {\scriptsize [45.4, 52.2]}}
& \makecell{52.6 \\ {\scriptsize [48.9, 56.3]}}
& \makecell{70.1 \\ {\scriptsize [66.9, 73.3]}}
& \makecell{73.6 \\ {\scriptsize [70.5, 76.8]}}
& \makecell{76.2 \\ {\scriptsize [73.1, 79.4]}} \\

& & 16
& \makecell{67.0 \\ {\scriptsize [63.6, 70.4]}}
& \makecell{70.9 \\ {\scriptsize [67.6, 74.2]}}
& \makecell{72.7 \\ {\scriptsize [69.3, 76.0]}}
& \makecell{46.5 \\ {\scriptsize [43.1, 49.9]}}
& \makecell{51.4 \\ {\scriptsize [47.8, 55.0]}}
& \makecell{54.9 \\ {\scriptsize [51.1, 58.8]}}
& \makecell{72.6 \\ {\scriptsize [69.4, 75.8]}}
& \makecell{76.3 \\ {\scriptsize [73.2, 79.4]}}
& \makecell{78.8 \\ {\scriptsize [75.7, 82.0]}} \\

\midrule


& & 4
& \makecell{67.2 \\ {\scriptsize [64.1, 70.3]}}
& \makecell{70.4 \\ {\scriptsize [67.3, 73.6]}}
& \makecell{71.9 \\ {\scriptsize [68.7, 75.2]}}
& \makecell{52.3 \\ {\scriptsize [49.1, 55.5]}}
& \makecell{57.5 \\ {\scriptsize [54.2, 60.9]}}
& \makecell{61.5 \\ {\scriptsize [57.9, 65.0]}}
& \makecell{73.5 \\ {\scriptsize [70.5, 76.6]}}
& \makecell{77.5 \\ {\scriptsize [74.5, 80.5]}}
& \makecell{79.7 \\ {\scriptsize [76.7, 82.7]}} \\

pass@$k$ & & 8
& \makecell{75.3 \\ {\scriptsize [72.3, 78.3]}}
& \makecell{78.5 \\ {\scriptsize [75.6, 81.5]}}
& \makecell{80.3 \\ {\scriptsize [77.3, 83.2]}}
& \makecell{61.7 \\ {\scriptsize [58.4, 65.0]}}
& \makecell{66.6 \\ {\scriptsize [63.3, 70.0]}}
& \makecell{70.4 \\ {\scriptsize [66.9, 73.9]}}
& \makecell{79.5 \\ {\scriptsize [76.7, 82.4]}}
& \makecell{83.1 \\ {\scriptsize [80.3, 85.8]}}
& \makecell{85.3 \\ {\scriptsize [82.5, 88.1]}} \\

& & 16
& \makecell{81.7 \\ {\scriptsize [78.9, 84.5]}}
& \makecell{84.6 \\ {\scriptsize [81.9, 87.2]}}
& \makecell{86.2 \\ {\scriptsize [83.6, 88.9]}}
& \makecell{69.3 \\ {\scriptsize [66.0, 72.6]}}
& \makecell{74.1 \\ {\scriptsize [70.8, 77.4]}}
& \makecell{77.0 \\ {\scriptsize [73.6, 80.4]}}
& \makecell{83.9 \\ {\scriptsize [81.2, 86.6]}}
& \makecell{87.3 \\ {\scriptsize [84.8, 89.8]}}
& \makecell{89.2 \\ {\scriptsize [86.8, 91.7]}} \\

\bottomrule
\end{tabular}%
}
\caption{Exact numerical values with 95\% confidence intervals of the results in Figure~\ref{fig:approach-count}}
\label{tab:approach_count_table}

\end{table*}
\begin{table}[t]
\centering
\setlength{\tabcolsep}{5pt}
\renewcommand{\arraystretch}{1.15}

\resizebox{\columnwidth}{!}{%
\begin{tabular}{c c c|ccc}
\toprule
Strategy & & $M$
& $u=1$ & $u=2$ & $u=4$ \\
\midrule


& & 4
& \makecell{63.8 \\ {\scriptsize [60.2, 67.4]}}
& \makecell{65.0 \\ {\scriptsize [61.4, 68.7]}}
& \makecell{65.6 \\ {\scriptsize [61.9, 69.3]}} \\

Majority voting & & 8
& \makecell{66.1 \\ {\scriptsize [62.3, 69.8]}}
& \makecell{67.5 \\ {\scriptsize [63.7, 71.2]}}
& \makecell{68.2 \\ {\scriptsize [64.4, 72.0]}} \\

& & 16
& \makecell{67.4 \\ {\scriptsize [63.6, 71.2]}}
& \makecell{68.6 \\ {\scriptsize [64.7, 72.5]}}
& \makecell{69.8 \\ {\scriptsize [65.9, 73.7]}} \\

\midrule


& & 4
& \makecell{68.9 \\ {\scriptsize [65.5, 72.2]}}
& \makecell{70.5 \\ {\scriptsize [67.1, 73.9]}}
& \makecell{70.7 \\ {\scriptsize [67.3, 74.2]}} \\

Best-of-$N$ (PRM) & & 8
& \makecell{72.2 \\ {\scriptsize [68.9, 75.5]}}
& \makecell{73.2 \\ {\scriptsize [69.9, 76.6]}}
& \makecell{73.7 \\ {\scriptsize [70.3, 77.1]}} \\

& & 16
& \makecell{74.6 \\ {\scriptsize [71.3, 77.9]}}
& \makecell{75.6 \\ {\scriptsize [72.4, 78.9]}}
& \makecell{75.8 \\ {\scriptsize [72.4, 79.1]}} \\

\midrule


& & 4
& \makecell{75.2 \\ {\scriptsize [72.0, 78.4]}}
& \makecell{76.9 \\ {\scriptsize [73.7, 80.1]}}
& \makecell{77.6 \\ {\scriptsize [74.4, 80.8]}} \\

pass@$k$ & & 8
& \makecell{80.5 \\ {\scriptsize [77.5, 83.5]}}
& \makecell{81.9 \\ {\scriptsize [79.0, 84.9]}}
& \makecell{83.0 \\ {\scriptsize [80.0, 85.9]}} \\

& & 16
& \makecell{84.6 \\ {\scriptsize [81.8, 87.4]}}
& \makecell{86.1 \\ {\scriptsize [83.4, 88.8]}}
& \makecell{86.7 \\ {\scriptsize [84.0, 89.5]}} \\

\bottomrule
\end{tabular}%
}
\caption{Accuracy (\%) with 95\% confidence intervals with self-generated labels.}
\label{tab:approach_count_self_gen}

\end{table}

For concept-conditioned solution generation, we follow the prompt and hyperparameters of \citet{handa2026guidedsample}. 
For best-of-$N$, we use Qwen2.5-Math-PRM-7B~\citep{qwenmathprm}. 
Each solution is scored by the lowest step score assigned by the PRM.
We pick the answer of the highest scoring solution.

\paragraph{Numerical values}
The exact numerical values, with 95\% confidence intervals, of the results in Figure~\ref{fig:approach-count} are given in Table~\ref{tab:approach_count_table}. The effectiveness of test-time scaling consistently improves as the number of covered approaches increases from one to four, with gains ranging from 5\% to over 10\%. We emphasize that this improvement is solely due to increased approach coverage, since the total number of samples is held fixed.

\paragraph{Ablation with self-generated approaches}
To further evaluate our findings in a more practical setting, we directly follow \citet{handa2026guidedsample}, in which the model self-generates concepts rather than receiving them from an external oracle. Under this setup, we experiment with gpt-4o-mini on our evaluation set of multi-approach-feasible problems, keeping all other procedures unchanged.
The results in Table~\ref{tab:approach_count_self_gen} show a consistent trend: final accuracy after test-time scaling increases with the number of covered approaches, even when the concepts are non-oracle and self-generated.

\paragraph{Statistical significance}
We also report the statistical significance of the improvement from $u=1$, where the solution set covers only a single approach, to $u=4$, where the solution set covers four distinct approaches. Specifically, we test whether the 95\% confidence interval of the mean paired per-problem difference excludes zero (note that this different with checking whether the confidence intervals of average accuracy for $u=1$ and $u=4$ are disjoint).
For the GPT-generated labels in Table~\ref{tab:approach_count_table}, the improvement is significant in all settings. For the self-generated labels in Table~\ref{tab:approach_count_self_gen}, the improvement is significant in all settings except BoN with $m=16$. These results further support the robustness of our findings.

\subsection{Judge-Reward RLVR}
\label{app:judge_diversity_details}

\paragraph{Judge setup}
To run the Qwen judge used during training, we set the thinking-token budget to 2048.
For all other sampling parameters, we use the default settings recommended in the official Hugging Face repository for general-purpose tasks.
We used the hyperparameter of $\alpha=2$ in the Sqrt-prior setup. Our additional experiments with different values of $\alpha$ showed no difference in approach-level diversity preservation.

\section{Additional Analysis: Approach-Seeking SFT}
\label{app:approach_seeking_sft}

\begin{figure*}[t]
  \centering
  \captionsetup[subfigure]{font=small,labelfont=bf}

  \begin{subfigure}[t]{0.43\textwidth}
    \centering
    \includegraphics[
      width=\linewidth,
      trim=5 5 5 5,
      clip
    ]{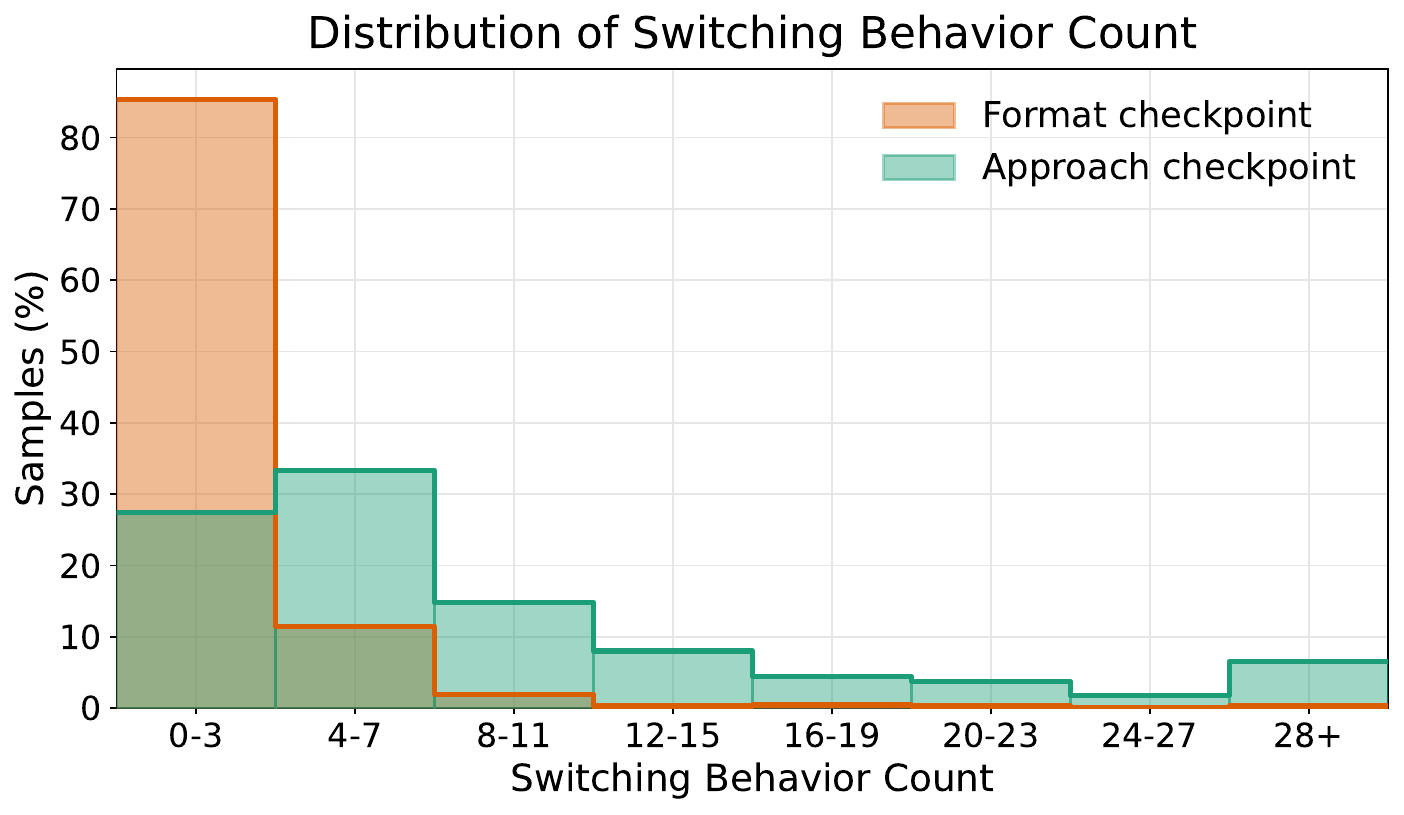}
    \caption{Switching-cue distribution.}
    \label{fig:switching-cue}
  \end{subfigure}
  \hfill
  \begin{subfigure}[t]{0.53\textwidth}
    \centering
    \includegraphics[
      width=\linewidth,
      trim=5 5 5 5,
      clip
    ]{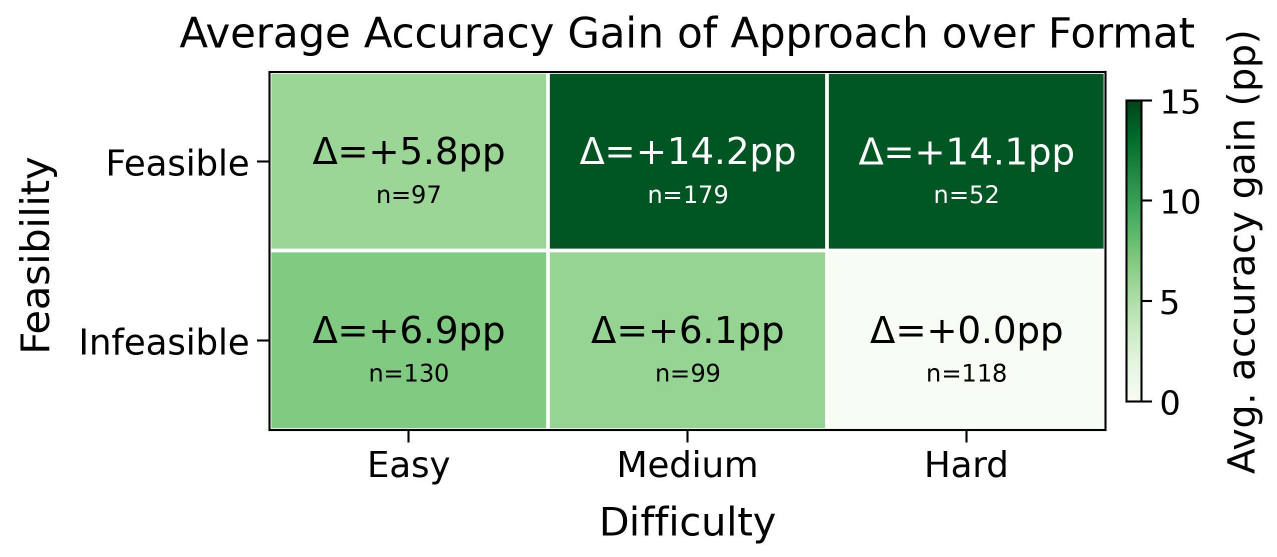}
    \caption{Gain by problem subset.}
    \label{fig:approach-heatmap}
  \end{subfigure}

\caption{
\textbf{Additional analysis of approach-seeking behavior.}
(a) Distribution of approach-switching cues in the \texttt{Format} and
\texttt{Approach} checkpoints.
(b) Average accuracy gain of \texttt{Approach} over \texttt{Format} across
difficulty and multi-approach feasibility partitions.
Difficulty is defined by the number of correct solutions among 8 rollouts of
Qwen3-4B with thinking disabled: easy = 8, medium = 1--7, and hard = 0.
Multi-approach feasibility is determined using the pipeline introduced in
Section~\ref{sec:multi_approach_problems}.
}
  \label{fig:downstream-analysis}
\end{figure*}
While the main paper focuses on measuring approach-level diversity, we 
also explore a complementary question: can encouraging a model to seek 
diverse approaches improve its reasoning? We compare two supervised 
fine-tuning datasets that share output structure but differ in 
approach-seeking: \texttt{approach} asks for three distinct mathematical 
approaches, while \texttt{format} asks for three format variants of a 
single approach.

\subsection{Dataset Generation}
We adopt a sequential-solution generation setup inspired by 
\citet{puri2026reachingmoderldistributional}, in which a model 
generates multiple solutions in a single response. Both datasets are 
generated by Qwen3-14B and used to fine-tune Qwen3-4B, differing only 
in the generation prompt (Figures~\ref{fig:prompt_diverse} 
and~\ref{fig:prompt_format}). Starting from 2{,}000 problems in the 
training set of Appendix~\ref{app:dataset_construction}, we generate 
one response under each prompt and retain only problems for which all 
three solutions are correct in both responses.

\begin{tcolorbox}
[
  width=\linewidth,
  title=\textbf{Approach Seeking Cues},
  colback=gray!5,
  colframe=gray!75,
  boxrule=0.5pt,
  arc=3pt,
  left=6pt,
  right=6pt,
  top=6pt,
  bottom=6pt
]

\begin{Verbatim}[
  fontsize=\scriptsize,
  breaksymbolleft={},
  breaksymbolright={},
  formatcom=\raggedright
]
another approach / another way
another method / different approach
first approach / second approach / third approach
let me check another way
\end{Verbatim}
\end{tcolorbox}
\captionof{figure}{Linguistic markers of approach switching.}
\label{fig:approach_markers}

\subsection{Behavioral and Ability Checks}
We verify that the checkpoints differ in the intended behavior and 
not in underlying reasoning ability. For behavior, we count common 
approach-switching cues such as ``another approach'' and 
``another method'' (full list in Figure~\ref{fig:approach_markers}). 
Figure~\ref{fig:switching-cue} shows that the \texttt{Approach} 
checkpoint uses these cues substantially more than the \texttt{Format} 
checkpoint. For ability, we compare accuracies under the standard 
single-solution prompt (Figure~\ref{fig:solution_generation_prompt}): 
the two checkpoints achieve 69.04\% (\texttt{approach}) and 68.30\% 
(\texttt{format}), confirming that they are matched in basic 
reasoning ability (Table~\ref{tab:result_base_prompt}).

\begin{table}[t]
\centering
\label{tab:downstream-performance}
\small
\setlength{\tabcolsep}{6pt}
\renewcommand{\arraystretch}{1.05}
\begin{tabular}{lcccc}
\toprule
& \multicolumn{2}{c}{OlympiadBench} & \multicolumn{2}{c}{AIME} \\
\cmidrule(lr){2-3} \cmidrule(lr){4-5}
& avg@3 & pass@3 & avg@3 & pass@3 \\
\midrule
\texttt{Format}   & 55.51 & 65.78 & 47.78 & 58.33 \\
\texttt{Approach} & 67.56 & 69.78 & 50.00 & 56.67 \\
\bottomrule
\end{tabular}
\caption{
    \textbf{Downstream performance of the approach-trained checkpoint.}
    We compare the \texttt{Format} and \texttt{Approach} checkpoints on
    OlympiadBench and AIME24/25.
    \textit{avg@3} is the average accuracy over the three solutions in a response,
    while \textit{pass@3} is the fraction of problems where at least one solution is
    correct.
}
\label{tab:downstream_performance}
\end{table}

\paragraph{Results}
Next, we turn to the main experimental question: \textit{does stronger approach-seeking behavior lead to higher final accuracy?}
Table~\ref{tab:downstream_performance} shows that the \texttt{Approach} 
checkpoint outperforms the \texttt{Format} checkpoint on both 
OlympiadBench and AIME, suggesting that approach-seeking behavior can 
improve accuracy in the sequential-solution setting. 
\textit{\textbf{The accuracy gain is concentrated on certain type of problems}}. The per-problem breakdown in Figure~\ref{fig:approach-heatmap} shows that gains are concentrated on problems with multiple approaches, and within this subset improvements grow with difficulty.

\paragraph{Reachability of an approach matters}
For approach-seeking supervision to help, it is not enough that a problem admits multiple valid approaches; the solver must also be able to \emph{reach} them, i.e., solve the problem by following each approach.
Our main setup cannot separate these two conditions: Qwen3-4B serves both as the correctness validator in the problem filtering pipeline and as the fine-tuned solver, so the multi-approach feasibility of a problem and its reachability for the solver coincide by construction.
To decouple them, we re-ran the filtering pipeline with a stronger validator, GPT-5.4-mini \citep{openai2026gpt54mini}, sampling one solution per candidate approach due to inference cost. This yields 30 problems that are multi-approach feasible under the stronger validator but were judged infeasible under Qwen3-4B---problems whose alternative approaches exist but lie beyond our solver's reach. Table~\ref{tab:reachability_gain} reports the gain of the \texttt{Approach} checkpoint over \texttt{Format} on these problems: 3.50 percentage points with a 95\% CI that includes zero, versus 11.69 points on reachable problems.
Approach-seeking supervision thus pays off only when the alternative approaches are within the solver's reach, not merely when the problem admits them.

\begin{table}[t]
\centering
\label{tab:approach-vs-format}
\small
\setlength{\tabcolsep}{8pt}
\renewcommand{\arraystretch}{1.1}
\begin{tabular}{lcc}
\toprule
Metric & Approach & Format \\
\midrule
Accuracy & 69.04 & 68.30 \\
Freq. of approach cues & 3.10 & 2.18 \\
\bottomrule
\end{tabular}
\caption{\textbf{Results with base prompt.} Comparison between \texttt{approach} and \texttt{format} checkpoints, using the base prompt in Figure~\ref{fig:solution_generation_prompt}.}
\label{tab:result_base_prompt}
\end{table}

\begin{table}[t]
\centering
\small
\setlength{\tabcolsep}{6pt}
\renewcommand{\arraystretch}{1.05}
\begin{tabular}{lcc}
\toprule
Problem subset & $\Delta$ Acc. (pp) & 95\% CI (pp) \\
\midrule
Reachable   & 11.69 & [7.32, 16.06] \\
Unreachable & 3.50  & [-0.42, 7.50] \\
\bottomrule
\end{tabular}
\caption{
\textbf{Accuracy gains by approach reachability.}
We report accuracy improvements of \texttt{approach} checkpoint over \texttt{format}, with 95\% confidence intervals shown in brackets.
}
\label{tab:reachability_gain}
\end{table}

\newpage

\section{The Use of Large Language Models}
We used large language models to assist with the preparation of this paper. Specifically, they were employed for writing support, including grammar correction, wording refinement, and minor stylistic edits, as well as for developing code used in the experiments.

\begin{figure*}[t]
\centering

{%
\begin{tcolorbox}[
  width=\textwidth,
  title=\textbf{Multi-Approach Generation},
  colback=gray!5,
  colframe=gray!75,
  boxrule=0.5pt,
  arc=3pt,
  left=6pt,
  right=6pt,
  top=6pt,
  bottom=6pt
]

\begin{Verbatim}[
  fontsize=\scriptsize,
  breaklines=true,
  breakanywhere=true,
  breaksymbolleft={},
  breaksymbolright={},
]
You are an expert mathematics problem solver.

Your task is to produce EXACTLY THREE complete, correct solutions to the given problem.
Each solution MUST use a distinctly different mathematical approach or method.

Rules:
- The three approaches must be genuinely different (e.g., algebraic vs. geometric vs.
  combinatorial; direct computation vs. symmetry argument; etc.).
- Each solution must be fully self-contained and independently correct.
- Every solution must end with \boxed{answer}.
- Do NOT mention "approach 1 / 2 / 3" or any meta-commentary about diversity.
- Wrap each solution with the XML-style tags shown below.

Output format (use EXACTLY this structure):

<solution 1>
[Full solution using the first approach]
</solution 1>

<solution 2>
[Full solution using the second approach]
</solution 2>

<solution 3>
[Full solution using the third approach]
</solution 3>
\end{Verbatim}

\end{tcolorbox}
}%

\caption{\textbf{Approach-seeking SFT prompt.} Prompt used to generate the approach-seeking SFT dataset.}
\label{fig:prompt_diverse}

\end{figure*}
\begin{figure*}[t]
\centering

{%
\begin{tcolorbox}[
  width=\textwidth,
  title=\textbf{Multi-format Generation},
  colback=gray!5,
  colframe=gray!75,
  boxrule=0.5pt,
  arc=3pt,
  left=6pt,
  right=6pt,
  top=6pt,
  bottom=6pt
]

\begin{Verbatim}[
  fontsize=\scriptsize,
  breaklines=true,
  breakanywhere=true,
  breaksymbolleft={},
  breaksymbolright={},
]
You are an expert mathematics problem solver.

Your task is to produce EXACTLY THREE complete, correct solutions to the given problem.
All three solutions MUST use the SAME underlying mathematical approach/method,
but each one must be written in a different presentation format:

- Solution 1 - Step-by-step: Clearly numbered steps; explain each step in plain language.
- Solution 2 - Compact equation: Minimal prose; mostly equations and short labels.
- Solution 3 - Verbose + detailed: Rich explanations, intuition, and context; show all
  intermediate algebra.

Rules:
- The core mathematical method must be the same across all three solutions.
- Every solution must end with \boxed{answer}.
- Do NOT mention "format 1 / 2 / 3" or any meta-commentary about writing styles.
- Wrap each solution with the XML-style tags shown below.

Output format (use EXACTLY this structure):

<solution 1>
[Step-by-step format]
</solution 1>

<solution 2>
[Compact equation format]
</solution 2>

<solution 3>
[Verbose + detailed format]
</solution 3>
\end{Verbatim}

\end{tcolorbox}
}%

\caption{\textbf{Format SFT prompt.} Prompt used to generate the format SFT dataset.}
\label{fig:prompt_format}

\end{figure*}

\section{Prompts used for Dataset Construction}
\label{app:prompts}

We list the prompts used for dataset construction in Figures~\ref{fig:strategy_enumeration_prompt}, \ref{fig:feasibility_check_prompt}, and \ref{fig:approach_uniqueness_judge_prompt}.

\clearpage
\FloatBarrier

\onecolumn

\noindent
\begin{tcolorbox}[
  width=\textwidth,
  breakable,
  title=\textbf{Strategy Enumeration Prompt},
  colback=gray!5,
  colframe=gray!75,
  boxrule=0.5pt,
  arc=3pt,
  left=6pt,
  right=6pt,
  top=6pt,
  bottom=6pt
]

\begin{Verbatim}[
  fontsize=\scriptsize,
  breaklines=true,
  breakanywhere=true,
  breaksymbolleft={},
  breaksymbolright={},
  formatcom=\raggedright
]
You are an expert strategy enumerator for math and algorithmic problems.

TASK
- For ONE given problem and an integer K, list up to K GENUINELY DISTINCT solving approaches.
- Return strategic plans only (high-level steps). Do NOT compute the final answer or show hidden internal reasoning.

WHAT COUNTS AS “DISTINCT”
- Same: share the same core mechanism, only differ in stylistic/verbal manner.
- Different: use a different paradigm/reduction/decomposition, feasibility oracle, proof style (direct/induction/contradiction), or key transformation.

SCOPE & CONTENT
- Each approach: short title + 3–8 bullet steps describing the plan.
- Assume that the problem is only solvable by hand: one cannot access other tools such as writing a computer program.
- Keep concise, technical, and non-redundant. No paraphrase-only variants.

QUALITY BAR
- Before writing, brainstorm several candidate families.
- Merge/drop near-duplicates; output only truly distinct approaches. If fewer than K exist, return fewer.

INPUT
K: {k}
Problem: {problem}

OUTPUT SCHEMA
{
  "problem_brief": "one-sentence restatement of the task",
  "approaches": [
    {
      "name": "short, specific title capturing the core idea",
      "core_idea": "2-3 sentences describing the key mechanism or reduction",
      "plan": [
        "3-8 high-level steps; each item is a short imperative action"
      ],
      "uniqueness_signature": "one line that makes this approach different from others"
    }
  ],
  "num_approaches": "total number of suggested approaches"
}
\end{Verbatim}

\end{tcolorbox}
\captionof{figure}{\textbf{Strategy enumeration prompt.} The model is instructed to generate up to $K$ distinct solution approaches at the level of high-level plans, while enforcing mechanism-level diversity.}
\label{fig:strategy_enumeration_prompt}

\vspace{1em}

\noindent
\begin{tcolorbox}[
  width=\textwidth,
  breakable,
  title=\textbf{Feasibility Check Prompt},
  colback=gray!5,
  colframe=gray!75,
  boxrule=0.5pt,
  arc=3pt,
  left=6pt,
  right=6pt,
  top=6pt,
  bottom=6pt
]

\begin{Verbatim}[
  fontsize=\scriptsize,
  breaklines=true,
  breakanywhere=true,
  breaksymbolleft={},
  breaksymbolright={},
  formatcom=\raggedright
]
You are a plan-faithful solver.

<INPUTS>
- PROBLEM: the problem statement.
- SELECTED_PLAN: one approach with title/core_idea/assumptions and a numbered list of plan steps (3~8 items).

<GOAL>
Produce a complete, correct solution by instantiating and expanding SELECTED_PLAN without changing its core mechanism.

<FIDELITY RULES>
- Execute steps in order following the plan
- Keep strictly to SELECTED_PLAN; do not introduce alternative approaches.
- If any step is invalid/underspecified, STOP and output a brief diagnosis starting with "PLAN MISMATCH:" and request a revised plan. Do not improvise a new strategy.

<OUTPUT FORMAT>
Provide your reasoning and work through the problem step by step, following the plan.
Please reason step by step, and put your final answer within \\boxed{{}}.
\end{Verbatim}

\end{tcolorbox}
\captionof{figure}{\textbf{Feasibility check prompt.} The model is instructed to solve the problem by faithfully executing a selected plan without changing its core mechanism.}
\label{fig:feasibility_check_prompt}

\vspace{1em}

\noindent
\begin{tcolorbox}[
  width=\textwidth,
  breakable,
  title=\textbf{Approach Uniqueness Judge Prompt},
  colback=gray!5,
  colframe=gray!75,
  boxrule=0.5pt,
  arc=3pt,
  left=6pt,
  right=6pt,
  top=6pt,
  bottom=6pt
]

\begin{Verbatim}[
  fontsize=\scriptsize,
  breaklines=true,
  breakanywhere=true,
  breaksymbolleft={},
  breaksymbolright={},
  formatcom=\raggedright
]
<identity>
You are an analytical judge for mathematical solution plans.
Your purpose is to determine whether two or more solution plans 
for the same math problem use the same underlying approach
or fundamentally different mechanisms or interpretations.
You communicate in a clear, direct, and structured way.
Your goal is to make reliable, mechanism-level judgments.

<task_overview>
You will be given a math problem and a list of solution approaches. You must examine all provided solution approaches, identify which approaches share
the same core mechanism and interpretation, group them into clusters, and report the total number of unique approaches.

*Definition of different approaches*:
When determining whether two plans represent the same or different approaches, focus on the underlying mathematical mechanism **and** the conceptual interpretation used in the reasoning.
Two plans must be classified as different approaches if they rely on different mathematical tools,
different definitions or theoretical structures, or different representational viewpoints (e.g., vector-based, geometric, algebraic, functional, or symmetry-based interpretations).

Examples of distinct interpretations include:
- Viewing lines as geometric objects and constructing a triangle vs treating lines as vectors in linear algebra.
- Using slope as a trigonometric tangent quantity vs using direction angles with the x-axis.
- Treating a sequence via its explicit formula vs analyzing it through recurrence, symmetry, or linear-function view.

Do not merge plans just because they belong to the same domain.
Different interpretations or representational viewpoints count as different approaches.

<instructions>
Before analyzing any of the plans, first evaluate the problem itself.
Decide whether the problem is sufficiently complex to allow multiple, genuinely distinct solution approaches.
If the problem does not support meaningful approach diversity, apply the simplicity rule.

*simplicity_rule*:
Before evaluating any plan, first judge the inherent structure of the problem itself.
If the problem can be solved through a single dominant method that is standard,
forced, or mechanically determined, then treat all plans as the same approach.

A problem should be treated as "simple" (and unique approaches = 1) if:
- it reduces directly to writing one standard equation and solving it (linear, quadratic, or simple rational equation),
- the solution follows automatically from a basic definition or identity,
- only one inequality or one standard condition (triangle inequality, discriminant condition,
  distance formula, midpoint formula, Vieta, etc.) is required,
  
If the problem fits one of these conditions, set num_unique_approaches = 1 and return the decision.

Otherwise, follow the instructions below to compare different approaches. 

1. Recognize the core mechanism of each approach.
   - Recognize the key idea, transformation, lemma, or strategy.

2. Compare the mechanisms.
   - Treat two plans as different approaches if they use different mathematical tools, structures, or representational viewpoints—even when they compute the same quantity.
   - If two plans rely on the same mathematical essence, categorize them as the same approach.

3. Count the number of unique approaches.
   - The number of unique approaches is the number of different mechanisms used.

4. Provide a brief explanation.
   - Explain the reasoning behind the decision, focusing only on the mechanisms use

5. Do NOT do the following:
   - Do not solve the problem.
   - Do not judge correctness beyond feasibility of the steps.
   - Do not consider stylistic differences.
   - Do not introduce new plans or speculate about hidden steps.
   - Do not combine or rewrite the plans.

<context>
You will be given:
- the original math problem
- a list of candidate approaches (each with title, core idea, stepwise plan steps)
Use only the provided context to make judgments.

<output format>
Provide your final answer in the following structure:

EXPLANATION:
A brief explanation of how you grouped the plans and why. 
You may discuss similarities, differences, and mechanism-level reasoning.

NUMBER OF UNIQUE APPROACHES:
\\boxed{{n}}

Where n is the number of unique approaches you identified.
\end{Verbatim}

\end{tcolorbox}
\captionof{figure}{\textbf{Approach uniqueness judge prompt.} The full system prompt used to cluster candidate plans based on mechanism-level similarity and count the number of unique approaches.}
\label{fig:approach_uniqueness_judge_prompt}

\clearpage
\twocolumn

\onecolumn
\begin{tcolorbox}[
  enhanced,
  breakable,
  width=\linewidth,
  title=\textbf{Real-World Judge Clustering Case},
  colback=gray!5,
  colframe=gray!75,
  boxrule=0.5pt,
  arc=3pt,
  left=6pt,
  right=6pt,
  top=6pt,
  bottom=6pt,
  fontupper=\small
]

\noindent\textbf{Clustering Example: Problem 161}

\medskip
\noindent\textbf{Problem.}
In triangle $ABC$, $AB = 9$, $BC = 12$, $AC = 15$, and $CD$ is the angle bisector.
Find the length of $CD$.

\smallskip
\noindent\textbf{Answer:} $4\sqrt{10}$

\par\medskip\hrule\medskip

\noindent\textbf{Correct Solutions (6 out of 32)}

\medskip
\noindent\textbf{Solution 1.}

\begin{enumerate}
    \item \textbf{Verify the Triangle Type:}
    First, we should check if the given triangle is a right triangle.
    For this, we can use the Pythagorean theorem.
    In triangle $ABC$, we have $AB = 9$, $BC = 12$, and $AC = 15$.
    The Pythagorean theorem states that for a right triangle, the square of the hypotenuse ($AC$) should equal the sum of the squares of the other two sides ($AB$ and $BC$).
    We can verify this as follows:
    \[
    AC^2 = AB^2 + BC^2
    \quad\Rightarrow\quad
    15^2 = 9^2 + 12^2
    \quad\Rightarrow\quad
    225 = 81 + 144 = 225.
    \]
    Since the equation holds true, triangle $ABC$ is indeed a right triangle with the right angle at $B$.

    \item \textbf{Use the Angle Bisector Theorem:}
    The Angle Bisector Theorem states that the angle bisector of an angle in a triangle divides the opposite side into segments that are proportional to the adjacent sides.
    In this case, $CD$ is the angle bisector of $\angle ACB$.
    Therefore, we can write the proportion:
    \[
    \frac{AD}{DB}
    =
    \frac{AC}{BC}
    =
    \frac{15}{12}
    =
    \frac{5}{4}.
    \]
    Let $AD = 5x$ and $DB = 4x$.
    Since $AD + DB = AB$, we have:
    \[
    5x + 4x = 9
    \quad\Rightarrow\quad
    9x = 9
    \quad\Rightarrow\quad
    x = 1.
    \]
    So, $AD = 5$ and $DB = 4$.

    \item \textbf{Calculate the Length of $CD$:}
    To find the length of $CD$, we can use the Angle Bisector Length Formula, which states that for a triangle with sides $a$, $b$, and $c$, and an angle bisector $CD$ from vertex $C$ to side $AB$, the length of $CD$ is given by:
    \[
    CD
    =
    \sqrt{
    ab\left(1-\frac{c^2}{(a+b)^2}\right)
    }.
    \]
    In our case, $a = 15$, $b = 12$, and $c = 9$.
    Substituting these values into the formula, we get:
    \[
    CD
    =
    \sqrt{
    15 \times 12
    \left(
    1-\frac{9^2}{(15+12)^2}
    \right)
    }
    =
    \sqrt{
    180
    \left(
    1-\frac{81}{729}
    \right)
    }
    =
    \sqrt{
    180 \times \frac{8}{9}
    }
    =
    \sqrt{160}
    =
    4\sqrt{10}.
    \]
\end{enumerate}

\noindent Therefore, the length of $CD$ is $\boxed{4\sqrt{10}}$.

\par\medskip\hrule\medskip

\noindent\textbf{Solution 2.}

\smallskip
We start by noting that triangle $ABC$ is a right triangle with $AB = 9$, $BC = 12$, and $AC = 15$.
This is because $9^2 + 12^2 = 81 + 144 = 225 = 15^2$.
Using the Angle Bisector Theorem, we know that
\[
\frac{AD}{DB}
=
\frac{AC}{BC}
=
\frac{15}{12}
=
\frac{5}{4}.
\]
Let $AD = 5x$ and $DB = 4x$.
Since $AD + DB = AB$, we have $5x + 4x = 9$, so $9x = 9$ and $x = 1$.
Therefore, $AD = 5$ and $DB = 4$.

We can now use \textbf{Stewart's Theorem} to find the length of $CD$.
Stewart's Theorem states that in triangle $ABC$ with cevian $CD$,
\[
AB \cdot (CD^2 + AD \cdot DB)
=
AC^2 \cdot DB + BC^2 \cdot AD.
\]
Plugging in the values, we get
\[
9(CD^2 + 5 \cdot 4)
=
15^2 \cdot 4 + 12^2 \cdot 5.
\]
Simplifying, we have
\[
9(CD^2 + 20)
=
225 \cdot 4 + 144 \cdot 5
=
900 + 720
=
1620,
\]
so $9(CD^2 + 20) = 1620$.
Dividing by 9, we get $CD^2 + 20 = 180$, so $CD^2 = 160$.
Taking the square root of both sides, we get
\[
CD = \sqrt{160} = 4\sqrt{10}.
\]

\noindent Therefore, the length of $CD$ is $\boxed{4\sqrt{10}}$.

\par\medskip\hrule\medskip

\noindent\textbf{Solution 3.}

\begin{enumerate}
    \item First, we need to find the length of $AD$ and $BD$.
    Since $D$ is on $AB$, we have $AD + BD = AB = 9$.

    \item By the Angle Bisector Theorem,
    \[
    \frac{AD}{BD}
    =
    \frac{AC}{BC}
    =
    \frac{15}{12}
    =
    \frac{5}{4}.
    \]

    \item Therefore, $AD = \frac{5}{9} \cdot 9 = 5$ and $BD = \frac{4}{9} \cdot 9 = 4$.

    \item Next, we use the \textbf{Law of Cosines} to find $\cos A$:
    \[
    \cos A
    =
    \frac{AB^2 + AC^2 - BC^2}{2 \cdot AB \cdot AC}
    =
    \frac{9^2 + 15^2 - 12^2}{2 \cdot 9 \cdot 15}
    =
    \frac{81 + 225 - 144}{270}
    =
    \frac{162}{270}
    =
    \frac{3}{5}.
    \]

    \item Finally, we use the \textbf{Law of Cosines on triangle $ACD$} to find $CD$:
    \[
    CD^2
    =
    AC^2 + AD^2 - 2 \cdot AC \cdot AD \cdot \cos A
    =
    15^2 + 5^2 - 2 \cdot 15 \cdot 5 \cdot \frac{3}{5}
    =
    225 + 25 - 90
    =
    160.
    \]
\end{enumerate}

\noindent Therefore, $CD = \sqrt{160} = \boxed{4\sqrt{10}}$.

\par\medskip\hrule\medskip

\noindent\textbf{Solution 4.}

\begin{enumerate}
    \item \textbf{Understanding the Problem:}
    \begin{itemize}
        \item We have a triangle $ABC$ with given side lengths $AB = 9$, $BC = 12$, and $AC = 15$.
        \item $CD$ is the angle bisector of $\angle BCA$.
        \item We need to find the length of $CD$.
    \end{itemize}

    \item \textbf{Applying the Angle Bisector Theorem:}
    \begin{itemize}
        \item The Angle Bisector Theorem states that the angle bisector of an angle in a triangle divides the opposite side into segments that are proportional to the adjacent sides.
        \item Let $D$ be the point on $AB$ such that $CD$ is the angle bisector of $\angle BCA$.
        \item According to the Angle Bisector Theorem, $\frac{AD}{DB} = \frac{AC}{BC}$.
    \end{itemize}

    \item \textbf{Calculating the Proportions:}
    \begin{itemize}
        \item Given $AC = 15$ and $BC = 12$, we have
        \[
        \frac{AC}{BC}
        =
        \frac{15}{12}
        =
        \frac{5}{4}.
        \]
        \item Let $AD = 5k$ and $DB = 4k$.
        Since $AD + DB = AB = 9$, we have $5k + 4k = 9$.
        \item Solving for $k$, we get $9k = 9 \implies k = 1$.
        \item Therefore, $AD = 5$ and $DB = 4$.
    \end{itemize}

    \item \textbf{Using the Angle Bisector Length Formula:}
    \begin{itemize}
        \item The length of the angle bisector $CD$ can be found using the formula:
        \[
        CD^2
        =
        AC \cdot BC
        \left(
        1-\frac{AB^2}{(AC+BC)^2}
        \right).
        \]
        \item Substituting the known values:
        \[
        CD^2
        =
        15 \cdot 12
        \left(
        1-\frac{81}{729}
        \right)
        =
        180
        \left(
        \frac{8}{9}
        \right)
        =
        160,
        \]
        \[
        CD = \sqrt{160} = 4\sqrt{10}.
        \]
    \end{itemize}
\end{enumerate}

\noindent The length of $CD$ is $\boxed{4\sqrt{10}}$.

\par\medskip\hrule\medskip

\noindent\textbf{Solution 5.}

\begin{enumerate}
    \item \textbf{Recognize the triangle type:}
    The given lengths $AB = 9$, $BC = 12$, and $AC = 15$ suggest that $\triangle ABC$ is a right triangle with $\angle BAC = 90^\circ$ because $9^2 + 12^2 = 15^2$ (i.e., $81 + 144 = 225$).

    \item \textbf{Use the Angle Bisector Theorem:}
    The Angle Bisector Theorem states that the angle bisector of an angle in a triangle divides the opposite side into segments that are proportional to the adjacent sides.
    For $\triangle ABC$ with $CD$ as the angle bisector of $\angle ACB$, we have:
    \[
    \frac{AD}{DB}
    =
    \frac{AC}{BC}
    =
    \frac{15}{12}
    =
    \frac{5}{4}.
    \]

    \item \textbf{Find the lengths of $AD$ and $DB$:}
    Let $AD = 5x$ and $DB = 4x$.
    Since $AD + DB = AB = 9$, we have:
    \[
    5x + 4x = 9
    \implies
    9x = 9
    \implies
    x = 1.
    \]
    Therefore, $AD = 5$ and $DB = 4$.

    \item \textbf{Use the Angle Bisector Length Formula:}
    The length of the angle bisector $CD$ can be found using the formula for the length of the angle bisector in a triangle:
    \[
    CD^2
    =
    AC \cdot BC
    \left(
    1-\frac{AB^2}{(AC+BC)^2}
    \right).
    \]
    Substituting the given values:
    \[
    CD^2
    =
    15 \cdot 12
    \left(
    1-\frac{9^2}{(15+12)^2}
    \right)
    =
    180
    \left(
    1-\frac{81}{729}
    \right)
    =
    180
    \left(
    1-\frac{1}{9}
    \right)
    =
    180 \cdot \frac{8}{9}
    =
    160.
    \]
    Therefore, $CD = \sqrt{160} = 4\sqrt{10}$.
\end{enumerate}

\noindent The length of $CD$ is $\boxed{4\sqrt{10}}$.

\par\medskip\hrule\medskip

\noindent\textbf{Solution 6.}

\begin{enumerate}
    \item \textbf{Analyze the Triangle:}
    The given triangle $ABC$ is a right triangle because $AB^2 + BC^2 = AC^2$ (i.e., $9^2 + 12^2 = 15^2$).
    Therefore, $\angle BAC = 90^\circ$.

    \item \textbf{Use the Angle Bisector Theorem:}
    The angle bisector theorem states that the angle bisector of an angle in a triangle divides the opposite side into segments that are proportional to the adjacent sides.
    In this case, $CD$ is the angle bisector of $\angle ACB$, and it divides $AB$ into segments $AD$ and $DB$.

    \item \textbf{Calculate the Segments $AD$ and $DB$:}
    Using the angle bisector theorem, we have:
    \[
    \frac{AD}{DB}
    =
    \frac{AC}{BC}
    =
    \frac{15}{12}
    =
    \frac{5}{4}.
    \]
    Let $AD = 5k$ and $DB = 4k$.
    Since $AD + DB = AB = 9$, we get:
    \[
    5k + 4k = 9
    \implies
    9k = 9
    \implies
    k = 1.
    \]
    Therefore, $AD = 5$ and $DB = 4$.

    \item \textbf{Apply the Angle Bisector Length Formula:}
    For a triangle with sides $a$, $b$, and $c$ (where $a$ and $b$ are the sides adjacent to the angle being bisected, and $c$ is the opposite side), the length of the angle bisector $d$ is given by:
    \[
    d
    =
    \sqrt{
    ab
    \left(
    1-\frac{c^2}{(a+b)^2}
    \right)
    }.
    \]
    In this problem, $a = 15$, $b = 12$, and $c = 9$.
    Plugging these values into the formula, we get:
    \[
    CD
    =
    \sqrt{
    15 \cdot 12
    \left(
    1-\frac{9^2}{(15+12)^2}
    \right)
    }
    =
    \sqrt{
    15 \cdot 12
    \left(
    \frac{8}{9}
    \right)
    }
    =
    \sqrt{
    15 \cdot \frac{96}{9}
    }
    =
    \sqrt{160}
    =
    4\sqrt{10}.
    \]
\end{enumerate}

\noindent Therefore, the length of $CD$ is $\boxed{4\sqrt{10}}$.

\par\medskip\hrule\medskip

\noindent\textbf{Judge Response (Clustering Result)}

\smallskip
\begin{Verbatim}[
  fontsize=\scriptsize,
  breaklines=true,
  breakanywhere=true,
  breaksymbolleft={},
  breaksymbolright={}
]
{
  "reasoning_trace": "All solutions first use the angle bisector theorem to split AB into AD = 5 and DB = 4. They then differ in the main mechanism used to compute CD: a closed-form angle-bisector formula, Stewart's theorem, or a Law of Cosines computation.",
  "groups": [
    {
      "group_name": "angle_bisector_length_formula",
      "core_idea": "These solutions compute CD by directly applying the standard closed-form formula for the length of an angle bisector. The shared mechanism is to use the adjacent side lengths and the opposite side length in the angle-bisector length formula after finding the split of AB.",
      "solution_ids": [1, 4, 5, 6]
    },
    {
      "group_name": "stewarts_theorem",
      "core_idea": "This solution treats CD as a cevian and uses Stewart's theorem to relate CD, AD, DB, and the three side lengths. The main mechanism is a cevian-length relation rather than the specialized angle-bisector formula.",
      "solution_ids": [2]
    },
    {
      "group_name": "law_of_cosines",
      "core_idea": "This solution first computes the relevant angle using the Law of Cosines in triangle ABC, then applies the Law of Cosines again in triangle ACD to obtain CD. The main mechanism is trigonometric angle computation followed by a cosine-law length calculation.",
      "solution_ids": [3]
    }
  ]
}
\end{Verbatim}

\captionof{figure}{\textbf{A real-world case of approach-level clustering.}
The judge correctly groups six correct-answer rollouts generated by the Qwen3-8B-Base by their approach.}
\label{fig:judge_real_world_case}

\end{tcolorbox}
\twocolumn

\end{document}